\documentclass{article}
\usepackage{microtype}
\usepackage{graphicx}
\usepackage{subcaption}
\usepackage{booktabs}
\usepackage{multirow}
\usepackage{amsmath,amssymb,amsthm}
\usepackage{mathtools}
\usepackage{bbm}
\usepackage{enumitem}
\usepackage{url}
\usepackage{hyperref}
\usepackage{algorithm}
\usepackage{algorithmic}
\usepackage{xcolor}
\usepackage{colortbl}
\usepackage{amsmath}
\usepackage{amssymb}
\usepackage[most]{tcolorbox}
\usepackage{tabularx}
\usepackage{pifont}
\usepackage{enumitem}

\newtcolorbox{MainCaseCard}[2][]{
    enhanced,
    boxrule=1pt,
    arc=4pt,
    colframe=black!70, 
    colback=white,
    fonttitle=\bfseries\sffamily,
    title={#2},
    colbacktitle=gray!15,
    coltitle=black,
    attach boxed title to top left={xshift=10pt, yshift=-10pt, yshifttext=-10pt},
    boxed title style={colframe=black!70, colback=gray!15, sharp corners},
    top=1pt,
    segmentation style={draw=black!20, solid, line width=0.5pt},
    lower separated=false,
    before skip=1.5em,
    after skip=1.5em,
    breakable=false,
    #1
}

\tcbset{
    compare_style/.style={
        enhanced, boxrule=0.5pt, arc=1pt, fonttitle=\tiny\bfseries,
        left=4pt, right=4pt, top=4pt, bottom=4pt, 
        before upper={
            \tiny
            \setlength{\abovedisplayskip}{0pt}
            \setlength{\belowdisplayskip}{1pt}
            \setlength{\abovedisplayshortskip}{0pt}
            \setlength{\belowdisplayshortskip}{1pt}
            \setlength{\jot}{0pt}
        }
    }
}

\usepackage[preprint]{icml2026}

\icmltitlerunning{Internalizing LLM Reasoning via Discovery and Replay of Latent Actions}

\begin{document}

\twocolumn[
\icmltitle{
Internalizing LLM Reasoning via Discovery and Replay of Latent Actions
}

\begin{icmlauthorlist}
\icmlauthor{Zhenning Shi\textsuperscript{*}}{a}
\icmlauthor{Yijia Zhu\textsuperscript{*}}{a}
\icmlauthor{Junhan Shi}{a}
\icmlauthor{Xun Zhang}{d}
\icmlauthor{Lei Wang}{a}
\icmlauthor{Congcong Miao}{b}
\end{icmlauthorlist}

\icmlaffiliation{a}{Tsinghua University}
\icmlaffiliation{b}{National University of Singapore}
\icmlaffiliation{d}{Fuzhou University}

\icmlcorrespondingauthor{Congcong Miao}{mccmai@163.com}

\vskip 0.3in
]

\printAffiliationsAndNotice{\icmlEqualContribution}

\begin{abstract}
The internalization of chain-of-thought processes into hidden states has emerged as a highly efficient paradigm for scaling test-time compute. However, existing activation steering methods rely on static control vectors that fail to adapt to the non-stationary evolution of complex reasoning tasks. To address this limitation, we propose STIR (\textbf{S}elf-Distilled \textbf{T}ools for \textbf{I}nternal \textbf{R}easoning), a framework that reformulates reasoning enhancement as a dynamic latent trajectory control problem.
STIR introduces a synergistic three-stage pipeline: (1) \textit{differential intrinsic action induction} harvests latent reasoning successes to crystallize steering primitives; (2) \textit{sparse control basis construction} curates a compact, geometrically diverse tool library; and (3) \textit{value-modulated trajectory intervention} dynamically injects context-specific impulses via anchor-based gating.
Extensive experiments on six arithmetic and logical benchmarks across four representative models demonstrate that STIR improves average accuracy by 1.9\% to 7.5\% while reducing average token consumption by up to 35\% compared to vanilla decoding. These findings demonstrate that the benefits of explicit chain-of-thought can be realized through dynamic latent trajectory control, internalizing the reasoning process to bypass the explicit generation while achieving superior fidelity.
Our code is available at \url{https://github.com/sznnzs/LLM-Latent-Action}.
\end{abstract}

\section{Introduction}
\label{sec:intro}

Large Language Models have achieved a paradigm shift in complex problem solving primarily through test-time compute scaling strategies~\cite{zhang2025survey}. While chain-of-thought prompting serializes reasoning into explicit tokens to tackle intricate challenges~\cite{zhou2022least}, other approaches, such as tree search~\cite{zhang2024rest} algorithms and majority voting~\cite{feng2025optimal}, further amplify performance by exploring diverse reasoning paths. However, these token centric paradigms impose a fundamental architectural constraint where the reasoning capability is rigidly coupled with the sequence length~\cite{zeng2025revisiting}. This linear dependency incurs substantial computational overhead and limits the scalability of deployment in resource constrained environments~\cite{sui2025stop}.

In response to these limitations, recent research has pivoted toward the concept of implicit reasoning, which aims to internalize the deductive process within the continuous hidden states of the model~\cite{chen2025reasoning, zhu2025scaling}. By compressing the inferential trajectory into the residual stream, this approach aims to decouple computational depth from surface form generation~\cite{hao2024training}. However, current implementations face significant hurdles. They predominantly rely on a fixed number of latent tokens, which fails to adapt to varying problem complexities. Furthermore, instilling this capability necessitates computationally expensive multi-stage training or extensive fine tuning. Critically, these methods struggle to scale to larger numbers of latent tokens as the lack of explicit scaffolding causes the hidden states to degenerate, leading to logic degradation and unreliable conclusions~\cite{su2025token, zhang2025latent}.

Activation steering offers a potential mechanism to exert control over these internal representations without retraining~\cite{turner2024activation, rimsky2024steering}. While early work focused on modulating global attributes using function vectors~\cite{todd2023function} or representation engineering~\cite{bartoszcze2025representation}, recent advancements have begun applying steering to reasoning tasks~\cite{chen2025seal}. These methods attempt to mitigate specific failure modes like overthinking~\cite{huang2025mitigating} or repetitive looping by injecting corrective vectors~\cite{azizi2025activation}. Despite their promise, they predominantly rely on static control vectors which apply a global bias uniformly across generations. We argue that this static paradigm is fundamentally ill-suited for complex reasoning. Reasoning is an intrinsically non-stationary process that evolves through distinct cognitive phases. A fixed steering pattern that suppresses verification might actively interfere with necessary error checking in subsequent steps, leading to semantic drift.
Consequently, there is a pressing need for dynamic control that can adapt to the evolution of the reasoning process.

To address these challenges, we propose STIR (\textbf{S}elf-Distilled \textbf{T}ools for \textbf{I}nternal \textbf{R}easoning), a framework that reformulates reasoning enhancement as a dynamic latent trajectory control problem. Our core insight is that the optimal control signals for rectifying reasoning errors are not external heuristics but are already latent within the exploration space of the model itself. Even when a model fails during greedy decoding, it often traverses correct subpaths or generates successful solutions in stochastic rollouts. STIR harvests these transient moments of reasoning success and crystallizes them into a library of reusable steering primitives. This approach transforms the fleeting internal dynamics of the model into a structured control mechanism that can be retrieved and applied on demand.

Specifically, the STIR framework operates through three synergistic stages to ensure precise intervention. First, to extract reliable control signals from generation traces, we employ \textit{differential intrinsic action induction}. By aligning stochastic rollouts at critical decision points, we compute the difference between the centroids of successful and failed latent states, isolating the implicit gradient required to rectify specific errors. Second, to manage the high dimensionality of the extracted vectors, we implement \textit{sparse control basis construction}. We utilize a geometric optimization objective that balances individual utility with collective orthogonality, selecting a compact basis that spans the topological manifold without redundancy. Finally, during inference, we deploy \textit{value modulated trajectory intervention} that acts as a dynamic controller. This module executes a retrieve-preview-commit cycle where candidate tools are validated via lookahead probing, and an anchor-based gating mechanism enforces abstention when the intrinsic trajectory is already optimal, preventing unnecessary perturbations.

We evaluate STIR on six benchmarks covering arithmetic reasoning and logical deduction tasks. The experimental results demonstrate that our framework establishes a new accuracy efficiency Pareto frontier. By effectively internalizing the benefits of search into the latent space, STIR improves average reasoning accuracy by 1.9\% to 7.5\% compared to standard chain-of-thought decoding while simultaneously reducing average token consumption by up to 35\%.
These findings suggest that the rigorous deliberation of system 2 search can be efficiently compiled into the latent dynamics of system 1, achieving high-fidelity reasoning.

Our contributions are summarized as follows:
\begin{itemize}
    \item We identify the temporal mismatch between static activation steering and dynamic reasoning processes and propose a control-centric formulation that treats reasoning enhancement as a trajectory rectification problem.
    \item We propose STIR, a framework that employs differential intrinsic action induction to distill latent tools from the contrastive rollouts and implements sparse control basis construction to build a diverse library.
    \item We design a robust runtime mechanism combining sparse retrieval with value-modulated trajectory intervention which allows the model to self-correct only when necessary via anchor-based gating.
    \item Extensive experiments on six arithmetic and logical benchmarks across four representative models demonstrate that STIR significantly outperforms baselines and achieves superior computational efficiency.
\end{itemize}

\begin{figure*}
    \centering
    \includegraphics[width=1\linewidth]{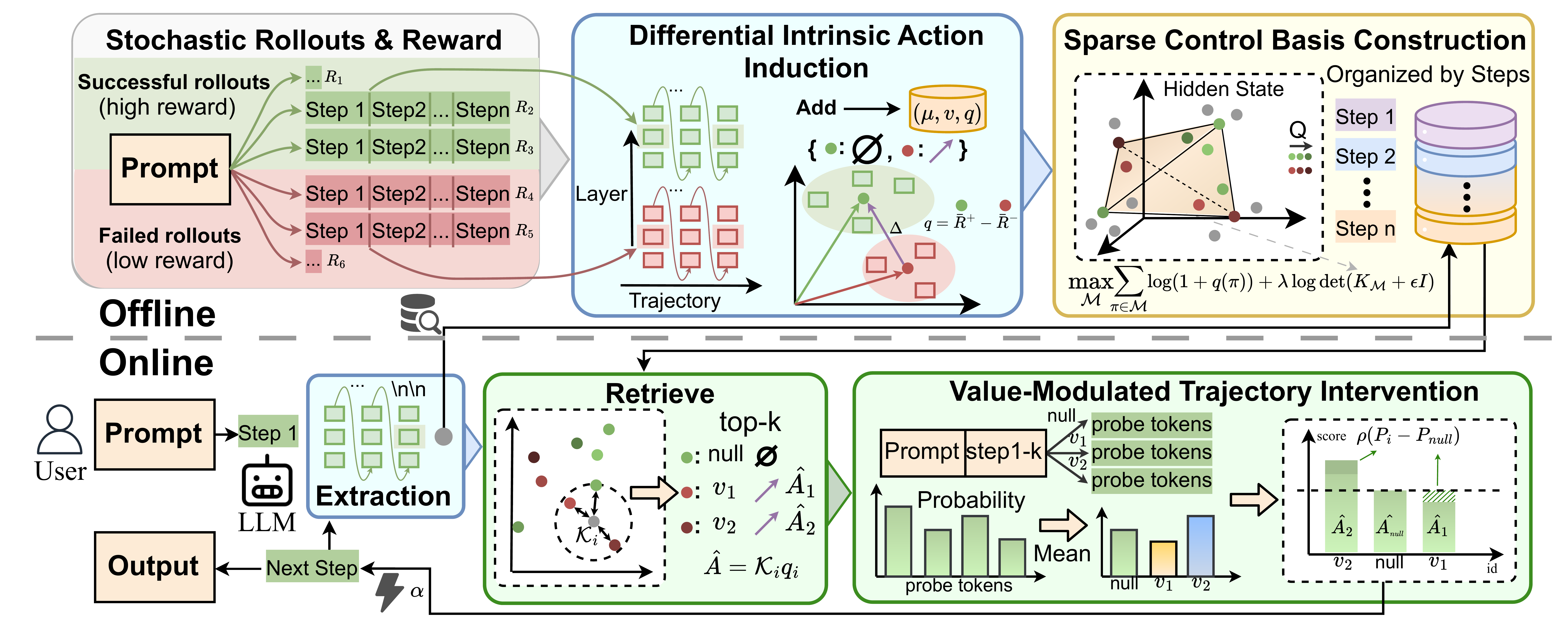}
    \caption{
Overview of the STIR framework. The pipeline operates through three stages: (1) differential intrinsic action induction distills latent steering impulses by analyzing the contrastive residuals between high-reward and low-reward rollouts at critical decision points; (2) sparse control basis construction filters these raw candidates into a geometrically diverse tool library to maximize representational coverage; and (3) value-modulated trajectory intervention acts as a runtime controller that retrieves relevant steering impulses and validates them via lookahead probing before dynamically injecting them into the residual stream.
}
    \label{fig:overview}
\end{figure*}

\section{Related Work}
\label{sec:related}

\paragraph{Large Language Models Reasoning.}
Current strategies for enhancing the reasoning capabilities of large language models predominantly focus on explicit token generation~\cite{zelikman2022star}. The foundational chain-of-thought prompting~\cite{wei2022chain} demonstrated that inducing intermediate reasoning steps significantly boosts performance on complex tasks~\cite{kojima2022large}. Subsequent research has structured this generation process into more elaborate topologies, such as tree-based search algorithms~\cite{yao2023tree} and graph-based reasoning networks~\cite{besta2024graph}, which allow for backtracking and lookahead. Parallel to these structural approaches, ensemble-based strategies leverage the diversity of stochastic sampling to marginalize out errors through majority voting~\cite{snell2024scaling}. While these methods effectively operationalize logical thinking, they create a linear dependency between reasoning depth and computational latency. Our work challenges this token-centric assumption, proposing that equivalent reasoning benefits can be achieved more efficiently by directly manipulating the latent trajectory.

\paragraph{Implicit Reasoning.}
Recent research has begun to investigate the extent to which reasoning can be internalized within the model's hidden states without explicit token output~\cite{hao2024training, zhu2025survey, shen2025codi}. Studies on implicit chain-of-thought suggest that transformer models can learn to perform complex deductions internally~\cite{zelikman2024quiet, skean2025layer}, compressing the reasoning process into the residual stream~\cite{lin2025implicit}. Similarly, works on pause tokens and internal computation tokens attempt to allocate additional latent capacity for thinking by extending the sequence with non-semantic placeholders~\cite{kim2025learning, london2025pause}. While these approaches demonstrate the plasticity of the latent space, they often lack structural constraints and precise control. They typically rely on opaque optimization procedures, making it difficult to actively rectify specific failures. STIR bridges the gap between explicit scaffolding and implicit computation by discretizing the continuous latent state evolution into retrievable, functional units.

\paragraph{Latent Representation Steering.}
The field of mechanistic interpretability has enabled a shift from prompt engineering to direct intervention within the model's internal representations~\cite{todd2023function}. Techniques such as activation steering~\cite{rimsky2024steering} and representation engineering~\cite{turner2023steering} have demonstrated that high-level concepts are linearly encoded within the activation space~\cite{stolfo2024improving}. By extracting a vector that represents a specific attribute~\cite{azizi2025activation}, these methods can control model behavior by injecting a fixed bias during inference~\cite{huang2025mitigating}. Despite their success in steering global attributes, these static intervention protocols face limitations. Reasoning is intrinsically a heterogeneous process involving distinct cognitive phases, including retrieval and logical deduction. A static control vector lacks the temporal flexibility to address these evolving requirements. Our framework advances this field by introducing a dynamic mechanism that adapts the intervention to the instantaneous state of the reasoning process.

\section{Preliminaries and Problem Formulation}
\label{sec:preliminaries}

\subsection{Latent Dynamics of Generative Reasoning}
\label{sec:prelim:dynamics}

We analyze the generative reasoning of a transformer-based large language model $\mathcal{M}_\theta$ as a discrete dynamical system. Given an input context $x$, the model generates a reasoning chain $Y = (y_1, \dots, y_T)$ where the hidden state $h_t^{(l)}$ at timestep $t$ and layer $l$ evolves via residual dynamics:
\begin{equation}
    h_t^{(l)} = h_t^{(l-1)} + \mathcal{F}_\theta^{(l)}(h_t^{(l-1)}, \text{Attn}(h_{<t})),
    \label{eq:residual_dynamics}
\end{equation}
Here, the final state $h_t^{(L)}$ projects onto the vocabulary to determine the next-token probability $P(y_t \mid x, y_{<t}) = \operatorname{softmax}(W_U h_t^{(L)})$. This residual structure implies that applying a targeted perturbation $\delta$ to $h_t^{(l)}$ can steer the generation trajectory without modifying the parameters $\theta$.

\subsection{The Challenge of Temporal Misalignment}
\label{sec:prelim:challenge}

Standard activation steering methods assume that a desired behavior corresponds to a global translation direction~\cite{chen2025persona, huang2025mitigating}, injecting a time-invariant vector $v^*$ into the residual stream:
\begin{equation}
    \tilde{h}_t \leftarrow h_t + \alpha \cdot v^*, \quad \forall t \in [1, T],
    \label{eq:static_steering}
\end{equation}
We identify a fundamental conflict when applying this static paradigm to multi-step reasoning. Reasoning is a heterogeneous process where optimal control directions vary across cognitive phases (e.g., hypothesis retrieval vs. logical deduction). A static vector $v^*$ approximates these diverse requirements with a single mean, creating a temporal misalignment: a direction that facilitates creative exploration may inherently contradict the constraints required for strict verification. This conflict necessitates a transition from static injection to dynamic, state-dependent control, where interventions are adapted to the evolving reasoning context.

\section{Methodology}
\label{sec:method}

We propose STIR (\textbf{S}elf-Distilled \textbf{T}ools for \textbf{I}nternal \textbf{R}easoning), which advances activation steering from static injection to a dynamic control paradigm. Our core insight is that the optimal corrective signals for complex reasoning are already latent within the model's exploration space and can be distilled into reusable tools. Rather than imposing a uniform external vector, STIR enables the model to self-correct by dynamically retrieving context-specific steering impulses from a sparse memory of its own reasoning successes. As illustrated in Figure~\ref{fig:overview}, this process operates through three integrated phases: (1) differential intrinsic action induction to distill latent steering impulses from contrastive rollouts; (2) sparse control basis construction to select a geometrically diverse set of tools; and (3) value-modulated trajectory intervention to apply the optimal tool during inference.

\subsection{Differential Intrinsic Action Induction}
\label{sec:method:induction}

Our core insight is that intrinsic signals within the model's own probability landscape can be effectively leveraged to guide reasoning correction. We propose a self-supervised pipeline to distill these transient reasoning successes into reusable control primitives by analyzing the contrastive dynamics of exploration. Since standard chain-of-thought generation often suffers from verbosity or circular logic, we enforce a constraint that prioritizes both correctness and conciseness. For each training prompt $x$, we sample a set of stochastic rollouts $\{Y^{(k)}\}_{k=1}^K$ and evaluate them using a length-regularized reward function:
\begin{equation}
    R(Y) = \mathbb{I}(Y \in \mathcal{Y}^*) - \eta \cdot \frac{\mathcal{L}(Y)}{L_{max}},
    \label{eq:reward_function}
\end{equation}
where $\mathbb{I}(\cdot)$ indicates correctness against ground truth, $\mathcal{L}(Y)$ denotes the generation length, and $\eta$ is a penalty coefficient. This objective effectively penalizes redundant computation, encouraging the discovery of concise reasoning paths that bypass inefficient loops while maintaining accuracy.

To extract latent steering impulses, we align the sampled trajectories at structural checkpoints (we use the double newline delimiter \texttt{\textbackslash n\textbackslash n}).
At a decision point $m$, we partition the rollout prefixes into a high-reward positive set $\mathcal{P}_m$ and a low-reward negative set $\mathcal{N}_m$. Operating under the hypothesis that the transition from failure to success can be locally approximated by a linear translation~\cite{park2023linear}, we compute the centroid hidden states for both sets:
\begin{equation}
    \mu_m^+ = \frac{1}{|\mathcal{P}_m|} \sum_{y \in \mathcal{P}_m} h_m(y), \quad \mu_m^- = \frac{1}{|\mathcal{N}_m|} \sum_{y \in \mathcal{N}_m} h_m(y),
    \label{eq:centroids}
\end{equation}
The candidate steering impulse is then derived as the counterfactual difference $v_m = \mu_m^+ - \mu_m^-$. Intuitively, this vector represents the implicit gradient required to transport an erroneous state, approximated by $\mu_m^-$, back to the optimal manifold region defined by $\mu_m^+$.

A persistent challenge in activation steering is the risk of unnecessary intervention, where applying a correction to an already optimal trajectory induces unintended semantic drift. To mitigate this, we structure the distilled repository into a dual-entry system that explicitly encodes both the policy for rectification and the evidence for abstention. Each memory unit comprises a correction entry and an anchor entry. The correction entry is keyed by the negative centroid $k^- = \mu_m^-$ and linked to the steering impulse $v_m$, serving as a functional directive to rectify failure modes. Conversely, the anchor entry is keyed by the positive centroid $k^+ = \mu_m^+$ and associated with a null impulse. By concurrently storing these null-action anchors, the framework enables the inference-time controller to systematically distinguish between states requiring correction and those already aligned with the optimal trajectory, thereby preventing over-intervention on valid reasoning chains.

\subsection{Sparse Control Basis Construction}
\label{sec:method:basis}

The differential induction phase yields a candidate set $\Omega$ that is inherently over-complete, often dominated by redundant variations of common failure modes. Directly deploying this raw set would incur prohibitive retrieval latency and dilute the control signal with repetitive directions. To address this, we operate under the hypothesis that diverse surface-level reasoning failures collapse onto a low-dimensional manifold within the residual stream~\cite{ansuini2019intrinsic, loaiza2024deep}. Our objective is to distill $\Omega$ into a compact control basis $\mathcal{M} \subset \Omega$ ($|\mathcal{M}| \le B$) that effectively spans the topology of failure modes while maximizing the intrinsic utility of the selected tools.

We formulate this selection task as a geometric optimization problem. Unlike standard formulations that couple quality and diversity into a single kernel, we adopt a decoupled objective to explicitly control the trade-off between individual efficacy and collective coverage. Let $K \in \mathbb{R}^{N \times N}$ be the similarity kernel where $K_{ij} = \langle k_i, k_j \rangle$ represents the cosine alignment between state keys, and let $q(\pi)$ denote the quality score of a tool derived from its historical reward gain. We seek to find the subset $\mathcal{M}$ that maximizes the following joint utility function:
\begin{equation}
    \mathcal{J}(\mathcal{M}) = \sum_{\pi \in \mathcal{M}} \log(1 + q(\pi)) + \lambda \log \det(K_{\mathcal{M}} + \epsilon I),
    \label{eq:dpp_objective}
\end{equation}
where $\lambda$ governs the penalty for redundancy and $\epsilon I$ ensures numerical stability. The first term encourages the selection of high-reward tools, while the second term, representing the log-volume of the subspace spanned by the keys, enforces geometric orthogonality among the selected entries.

Since maximizing this objective is NP-hard, we employ a greedy approximation algorithm with submodular guarantees~\cite{gillenwater2012near}. In each iteration, we select the next tool $\pi^*$ to add to the current basis $\mathcal{M}_{t-1}$ by maximizing the marginal gain:
\begin{equation}
    \pi^* = \operatorname*{arg\,max}_{\pi \in \Omega \setminus \mathcal{M}_{t-1}} \left( \log(1 + q(\pi)) + \lambda \log \sigma^2_{\pi | \mathcal{M}_{t-1}} \right),
    \label{eq:greedy_selection}
\end{equation}
Here, $\sigma^2_{\pi | \mathcal{M}_{t-1}}$ represents the conditional variance of the candidate's key vector given the span of the current basis, quantifying the geometric innovation of the new entry. This formulation aligns with the code implementation, which efficiently computes these updates using incremental projections. Once the basis is established, we optimize it for low-latency online retrieval by projecting the consolidated keys onto the unit hypersphere, ensuring that the subsequent retrieval operates strictly on angular distance and remains robust to norm variations across different layers.

\subsection{Value-Modulated Trajectory Intervention}
\label{sec:method:online}

During the inference phase, the framework operates as a dynamic controller that continuously monitors the latent reasoning trajectory. At each critical decision point $t$, the system evaluates whether to intervene, abstain, or explore further to resolve ambiguity. We operationalize this control process through a retrieve-preview-commit cycle. To establish a prior distribution over the latent action space, we query the sparse control basis $\mathcal{M}$ using the current hidden state $h_t$. This identifies a local candidate set $\mathcal{C}_t$ through adaptive filtering, where each tool $\pi$ is evaluated by its similarity-weighted quality $\hat{A}(\pi) = \mathcal{K}(h_t, k_\pi) \cdot q(\pi)$. Here, $\mathcal{K}$ denotes the cosine similarity kernel and $q(\pi)$ represents the historical reward gain. Crucially, the repository includes both correction entries with active steering impulses and anchor entries associated with null actions. When the hidden state aligns closely with an anchor entry, suggesting the trajectory is already within a high-reward manifold, the system prioritizes the null action to preserve intrinsic stability.

As static retrieval similarity can be imprecise in high-dimensional spaces, we validate candidate utility using a short-horizon counterfactual probe. For the top candidates, we simulate a lightweight forward pass of $T_{probe}$ tokens to generate a transient trajectory. We quantify the dynamic gain $\mathcal{G}(\pi)$ as the improvement in model confidence relative to the unperturbed baseline:
\begin{equation}
\begin{aligned}
    \mathcal{G}(\pi) = \frac{1}{T_{probe}} \sum_{j=1}^{T_{probe}} \Big( &\log P_\theta(\tilde{y}_{t+j} \mid \tilde{h}_{\pi}) \\
    &- \log P_\theta(y_{t+j} \mid h_{default}) \Big),
\end{aligned}
    \label{eq:probe_gain}
\end{equation}
This probing step serves as a local estimator of the reward gradient and filters out tools that are semantically similar but functionally ineffective. To determine the final routing decision, we fuse the static prior and dynamic evidence into a unified utility score $S(\pi) = \beta \cdot \hat{A}(\pi) + \rho \cdot \mathcal{G}(\pi)$, where weights $\beta$ and $\rho$ balance historical reliability with instance-specific adaptability. The controller selects the action $\pi^*$ that maximizes this score. To prevent over-intervention, we enforce a hard abstention threshold $\tau_{null}$. If the utility of the best tool falls below this floor, the system defaults to the null action. Conversely, if an active tool is selected, the injection strength $\alpha$ is dynamically modulated by the decision confidence:
\begin{equation}
    h_t \leftarrow h_t + \text{clip}(k_{scale} \cdot S(\pi^*), 0, \alpha_{max}) \cdot v_{\pi^*}.
    \label{eq:adaptive_strength}
\end{equation}
This adaptive commitment mechanism ensures that interventions remain proportional to certainty. It applies subtle corrections for ambiguous deviations while reserving steering where retrieval and probing signals align strongly.

\section{Experiments}
\label{sec:experiments}

\subsection{Experimental Setup}
\label{sec:exp:setup}

\paragraph{Datasets and Target Models.}
To systematically evaluate reasoning fidelity across varying horizon lengths, we conduct experiments on six diverse benchmarks. For mathematical deduction, we utilize AIME 2024/2025~\cite{aime} and AMC 2023~\cite{amc2023} alongside MATH-500~\cite{hendrycksmath2021}. For knowledge-intensive reasoning, we employ ARC-Challenge~\cite{clark2018think} and OpenBookQA~\cite{mihaylov2018can}. Our evaluation involves two distinct open-weight model families: the distilled reasoning specialists DeepSeek-R1-Distill-Qwen (1.5B, 7B)~\cite{guo2025deepseek} and the general-purpose Qwen2.5-Instruct (3B, 7B)~\cite{yang2025qwen3}, ensuring the framework's scalability across different model architectures.

\paragraph{Baselines.}
We benchmark our framework against a comprehensive set of inference paradigms to isolate the benefits of dynamic internal routing. We establish intrinsic model capabilities using Vanilla and assess structure-driven prompting via Self-Discover~\cite{zhou2024self}. To quantify the impact of compute scaling, we compare against Self-Consistency~\cite{wang2022self, chen2024more}. Finally, we position STIR against advanced test-time control baselines, including DEER~\cite{yang2025dynamic}, a dynamic early-exit framework, and SEAL~\cite{chen2025seal}, a representative static activation steering method.

\paragraph{Implementation Details.}
All implementations are built upon the Easysteer~\cite{xu2025easysteer} library, utilizing the vLLM~\cite{kwon2023efficient} backend for high-throughput inference. Experiments were conducted on a server equipped with eight 80GB NVIDIA H100 GPUs. In the offline phase, we construct the control basis via a Quality-Diversity Determinantal Point Process (QD-DPP)~\cite{kulesza2012determinantal} to maximize geometric coverage. We report accuracy and token cost as primary metrics. Detailed hyperparameter configurations are provided in Appendix~\ref{subsec:hyperparameters}.

\subsection{Main Results}
\label{sec:main_results}

\definecolor{gcolor}{RGB}{0,130,0}
\definecolor{rcolor}{RGB}{180,0,0}
\definecolor{bcolor}{RGB}{0,0,180}

\newcommand{\res}[2]{#1 \textcolor{gray}{(#2)}}

\newcommand{\inc}[1]{\textcolor{gcolor}{+#1}}
\newcommand{\dec}[1]{\textcolor{rcolor}{-#1}}
\newcommand{\neu}[1]{\textcolor{gray}{+#1}}

\newcommand{\avgres}[4]{#1 \scriptsize{(#2)} / \textcolor{gray}{#3} \scriptsize{(\textbf{#4})}}

\newcommand{\overhead}[1]{\textcolor{rcolor}{$\uparrow$#1}}
\newcommand{\plus}[1]{\textcolor{rcolor}{#1}}
\newcommand{\saving}[1]{\textcolor{bcolor}{$\downarrow$#1}}

\newcommand{\baseavg}[2]{#1 \scriptsize{\textcolor{gray}{(--)}} / \textcolor{gray}{#2} \scriptsize{(1.0x)}}

\begin{table*}[t]
\centering
\scriptsize
\setlength{\tabcolsep}{5.0pt}
\renewcommand{\arraystretch}{1.1}

\caption{Main Results across 6 Datasets and 4 Models. 
Values represent Accuracy\% (Avg. Tokens). 
The final column summarizes the Average Accuracy (Change) and Avg. Cost (Reduction Rate) relative to Vanilla. 
Bold denotes the highest accuracy per block.
Notably, STIR establishes a superior Pareto frontier, matching or exceeding the accuracy of Self-Consistency while reducing token consumption by up to 35\%, significantly outperforming static control baselines.}

\label{tab:big_results}

\resizebox{\textwidth}{!}{
\begin{tabular}{llccccccc}
\toprule
\multirow{2}{*}{\textbf{Model}} & \multirow{2}{*}{\textbf{Method}} & \multicolumn{4}{c}{\textbf{Math Reasoning}} & \multicolumn{2}{c}{\textbf{General QA}} & \textbf{Average Stats} \\
\cmidrule(lr){3-6} \cmidrule(lr){7-8} \cmidrule(lr){9-9}
& & \textbf{AIME 24} & \textbf{AIME 25} & \textbf{AMC 23} & \textbf{MATH500} & \textbf{ARC-C} & \textbf{OBQA} & \textbf{Acc ($\Delta$) / Cost (Drop\%)} \\
\midrule
& Vanilla & \res{\textbf{40.0}}{11,282} & \res{30.0}{11,642} & \res{60.0}{7,103} & \res{72.0}{4,688} & \res{\textbf{59.2}}{652} & \res{50.0}{668} & \baseavg{51.9}{6,006} \\
& Self-Consistency & \res{35.0}{32,761} & \res{35.0}{28,535} & \res{\textbf{73.3}}{17,529} & \res{\textbf{84.0}}{9,441} & \res{54.5}{2,065} & \res{\textbf{53.8}}{2,089} & \avgres{\textbf{55.9}}{\inc{4.1}}{15,403}{\plus{x2.6}} \\
& Self-Discover & \res{20.0}{10,837} & \res{20.0}{9,884} & \res{50.0}{7,955} & \res{74.0}{3,979} & \res{57.2}{650} & \res{47.2}{662} & \avgres{44.7}{\dec{7.1}}{5,661}{\saving{5\%}} \\
& DEER & \res{30.0}{9,935} & \res{\textbf{40.0}}{9,208} & \res{66.7}{6,509} & \res{71.0}{2,364} & \res{56.9}{625} & \res{51.2}{656} & \avgres{52.6}{\inc{0.8}}{4,883}{\saving{18\%}} \\
& SEAL & \res{30.0}{11,802} & \res{20.0}{13,262} & \res{60.0}{7,495} & \res{62.0}{5,458} & \res{49.0}{665} & \res{50.0}{640} & \avgres{45.2}{\dec{6.7}}{6,554}{\overhead{9\%}} \\
\rowcolor{blue!5}  & STIR$_{k_{scale}=1.0}$ & \res{25.0}{11,265} & \res{30.0}{11,031} & \res{70.0}{6,835} & \res{71.5}{3,282} & \res{49.8}{444} & \res{47.8}{446} & \avgres{49.0}{\dec{2.8}}{5,550}{\saving{7\%}} \\
\rowcolor{blue!5} \multirow{-7}{*}{\shortstack[l]{\textbf{DeepSeek-R1}\\\textbf{Distill-Qwen}\\\textbf{1.5B}}} & STIR$_{k_{scale}=0.75}$ & \res{35.0}{11,930} & \res{35.0}{11,383} & \res{70.0}{6,645} & \res{74.8}{4,297} & \res{56.2}{669} & \res{51.4}{629} & \avgres{53.7}{\inc{1.9}}{5,926}{\saving{1\%}} \\
\midrule
& Vanilla & \res{55.0}{11,240} & \res{45.0}{9,622} & \res{86.7}{5,009} & \res{85.2}{2,760} & \res{77.9}{604} & \res{75.4}{580} & \baseavg{70.9}{4,969} \\
& Self-Consistency & \res{65.0}{27,942} & \res{45.0}{29,249} & \res{90.0}{14,587} & \res{\textbf{89.8}}{7,974} & \res{79.9}{1,889} & \res{\textbf{77.4}}{1,905} & \avgres{74.5}{\inc{3.6}}{13,924}{\plus{x2.8}} \\
& Self-Discover & \res{40.0}{9,548} & \res{45.0}{9,971} & \res{80.0}{3,788} & \res{84.5}{2,133} & \res{76.3}{520} & \res{73.8}{527} & \avgres{66.6}{\dec{4.3}}{4,415}{\saving{11\%}} \\
& DEER & \res{70.0}{9,560} & \res{50.0}{9,070} & \res{83.3}{4,327} & \res{80.8}{2,069} & \res{77.3}{572} & \res{74.0}{550} & \avgres{72.6}{\inc{1.7}}{4,358}{\saving{12\%}} \\
& SEAL & \res{60.0}{8,902} & \res{50.0}{9,417} & \res{90.0}{3,728} & \res{85.0}{3,189} & \res{77.0}{679} & \res{74.0}{643} & \avgres{72.7}{\inc{1.8}}{4,426}{\saving{10\%}} \\
\rowcolor{blue!5}  & STIR$_{k_{scale}=1.0}$ & \res{65.0}{9,151} & \res{50.0}{9,288} & \res{90.0}{4,148} & \res{87.0}{2,732} & \res{79.9}{534} & \res{73.2}{374} & \avgres{74.2}{\inc{3.3}}{4,371}{\saving{12\%}} \\
\rowcolor{blue!5} \multirow{-7}{*}{\shortstack[l]{\textbf{DeepSeek-R1}\\\textbf{Distill-Qwen}\\\textbf{7B}}} & STIR$_{k_{scale}=0.75}$ & \res{\textbf{65.0}}{8,941} & \res{\textbf{60.0}}{9,653} & \res{\textbf{93.3}}{4,000} & \res{88.0}{2,638} & \res{\textbf{82.9}}{600} & \res{76.8}{591} & \avgres{\textbf{77.7}}{\inc{6.8}}{4,404}{\saving{11\%}} \\
\midrule
& Vanilla & \res{5.0}{2,749} & \res{5.0}{2,559} & \res{43.3}{1,416} & \res{65.8}{872} & \res{80.9}{297} & \res{75.4}{263} & \baseavg{45.9}{1,359} \\
& Self-Consistency & \res{10.0}{2,991} & \res{0.0}{3,090} & \res{36.7}{2,796} & \res{63.2}{1,918} & \res{83.9}{906} & \res{\textbf{77.6}}{804} & \avgres{45.2}{\dec{0.7}}{2,084}{\overhead{53\%}} \\
& Self-Discover & \res{10.0}{2,497} & \res{5.0}{1,021} & \res{56.7}{1,446} & \res{62.7}{873} & \res{73.9}{323} & \res{70.0}{289} & \avgres{46.4}{\inc{0.5}}{1,075}{\saving{20\%}} \\
& DEER & \res{10.0}{2,084} & \res{0.0}{2,186} & \res{50.0}{1,566} & \res{61.0}{1,146} & \res{75.9}{354} & \res{69.2}{329} & \avgres{44.4}{\dec{1.5}}{1,278}{\saving{6\%}} \\
& SEAL & \res{0.0}{2,802} & \res{0.0}{2,571} & \res{50.0}{853} & \res{\textbf{70.0}}{1,002} & \res{\textbf{85.0}}{313} & \res{66.0}{276} & \avgres{45.2}{\dec{0.7}}{1,303}{\saving{4\%}} \\
\rowcolor{blue!5}  & STIR$_{k_{scale}=1.0}$ & \res{10.0}{1,509} & \res{0.0}{981} & \res{43.3}{875} & \res{65.8}{752} & \res{75.9}{265} & \res{76.4}{256} & \avgres{45.2}{\dec{0.7}}{773}{\saving{43\%}} \\
\rowcolor{blue!5} \multirow{-7}{*}{\shortstack[l]{\textbf{Qwen2.5}\\\textbf{3B-Instruct}}} & STIR$_{k_{scale}=0.75}$ & \res{\textbf{20.0}}{1,264} & \res{\textbf{15.0}}{1,023} & \res{\textbf{56.7}}{1,432} & \res{67.5}{984} & \res{84.3}{287} & \res{77.2}{259} & \avgres{\textbf{53.4}}{\inc{7.5}}{875}{\saving{35\%}} \\
\midrule
& Vanilla & \res{20.0}{1,305} & \res{10.0}{1,045} & \res{53.3}{1,853} & \res{74.5}{731} & \res{88.0}{304} & \res{78.2}{279} & \baseavg{54.0}{920} \\
& Self-Consistency & \res{20.0}{2,998} & \res{\textbf{25.0}}{2,983} & \res{43.3}{2,828} & \res{75.5}{1,919} & \res{88.6}{925} & \res{79.4}{844} & \avgres{55.3}{\inc{1.3}}{2,083}{\plus{x2.3}} \\
& Self-Discover & \res{15.0}{1,865} & \res{15.0}{878} & \res{60.0}{821} & \res{71.0}{709} & \res{79.6}{326} & \res{73.6}{280} & \avgres{52.4}{\dec{1.6}}{813}{\saving{11\%}} \\
& DEER & \res{20.0}{1,315} & \res{15.0}{986} & \res{50.0}{935} & \res{73.0}{822} & \res{88.3}{321} & \res{77.8}{294} & \avgres{54.0}{\inc{0.0}}{779}{\saving{15\%}} \\
& SEAL & \res{20.0}{992} & \res{10.0}{1,078} & \res{63.3}{672} & \res{\textbf{80.0}}{717} & \res{86.0}{320} & \res{79.0}{279} & \avgres{56.4}{\inc{2.4}}{676}{\saving{26\%}} \\
\rowcolor{blue!5}  & STIR$_{k_{scale}=1.0}$ & \res{20.0}{965} & \res{5.0}{911} & \res{56.7}{797} & \res{70.0}{622} & \res{88.0}{299} & \res{76.8}{239} & \avgres{52.7}{\dec{1.3}}{639}{\saving{30\%}} \\
\rowcolor{blue!5} \multirow{-7}{*}{\shortstack[l]{\textbf{Qwen2.5}\\\textbf{7B-Instruct}}} & STIR$_{k_{scale}=0.75}$ & \res{\textbf{25.0}}{1,096} & \res{20.0}{937} & \res{\textbf{66.7}}{888} & \res{75.5}{778} & \res{\textbf{89.0}}{305} & \res{\textbf{80.4}}{254} & \avgres{\textbf{59.4}}{\inc{5.4}}{710}{\saving{22\%}} \\
\bottomrule
\end{tabular}%
}
\end{table*}

Table~\ref{tab:big_results} presents a comprehensive quantitative evaluation across mathematical and general reasoning benchmarks, where STIR establishes a superior accuracy-efficiency Pareto frontier. Specifically, on rigorous math tasks such as AIME 24 and MATH500, static steering methods struggle to accommodate shifting cognitive contexts, often resulting in marginal improvements or even regression due to the temporal misalignment of control vectors. In contrast, STIR$_{k_{scale}=0.75}$ achieves a new state-of-the-art by dynamically injecting context-aware interventions. For instance, on the DeepSeek-R1-7B model, it improves average accuracy by +6.8\% relative to the vanilla baseline, significantly surpassing SEAL (+1.8\%) and DEER (+1.7\%). Furthermore, while compression methods frequently compromise accuracy for speed (e.g., -7.1\% for Self-Discover on DeepSeek-1.5B), our efficiency-focused variant STIR$_{k_{scale}=1.0}$ effectively streamlines redundant reasoning steps, reducing token consumption by 30\% on Qwen2.5-7B while maintaining comparable fidelity.
Crucially, STIR outperforms Self-Consistency on average accuracy (e.g., 77.7\% vs 74.5\% on DeepSeek-7B), proving that the error-correcting benefits of ensemble sampling can be effectively compressed into a single latent trajectory.
This demonstrates that high-quality inference can emerge from precise latent adjustments.

\subsection{Geometry of Contrastive Latent States}
\label{sec:mechanistic}

\begin{figure}[t]
    \centering
    \begin{subfigure}{0.49\columnwidth}
        \centering
        \includegraphics[width=\linewidth]{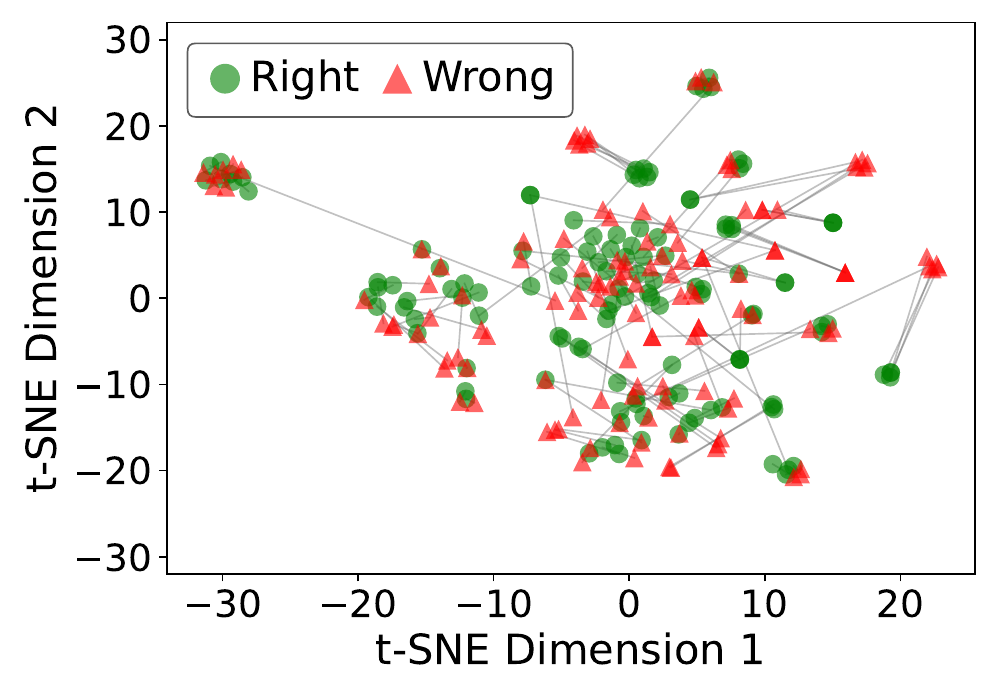}
        \caption{\textbf{MATH500}}
        \label{fig:tsne_a}
    \end{subfigure}
    \hfill
    \begin{subfigure}{0.49\columnwidth}
        \centering
        \includegraphics[width=\linewidth]{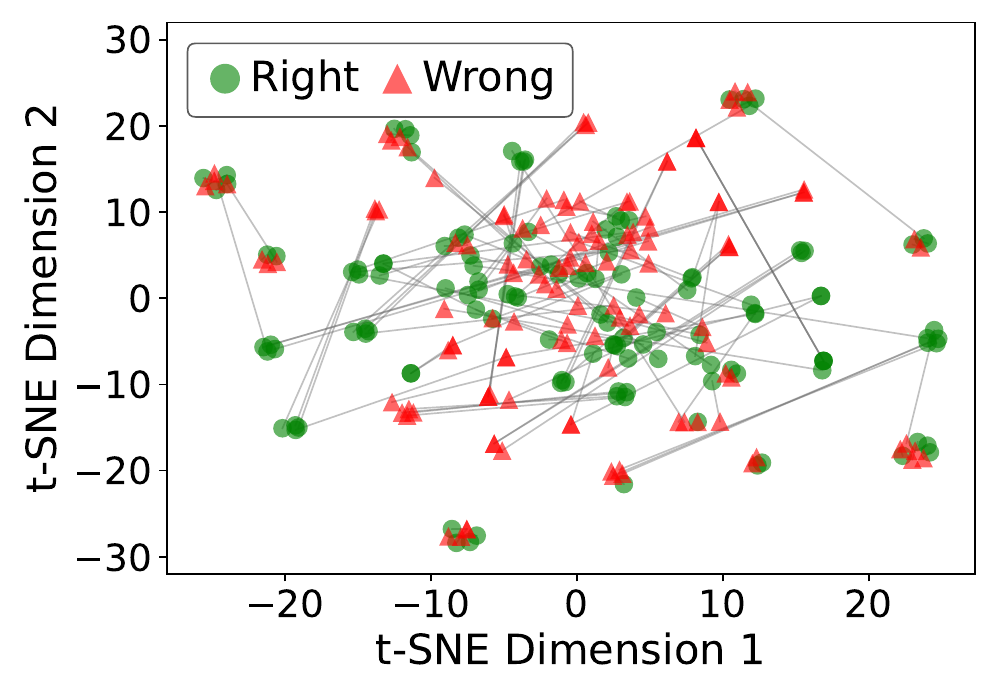}
        \caption{\textbf{OBQA}}
        \label{fig:tsne_b}
    \end{subfigure}
    
    \vspace{-0.2cm}
    \caption{t-SNE visualization of latent state embeddings extracted from stochastic rollouts. \textcolor{red}{$\blacktriangle$} denotes failure states ($\mu^-$), while \textcolor{gcolor}{$\bullet$} represents the corresponding rectified states ($\mu^+$). The directed edges illustrate the steering impulses, revealing that error states form geometrically coherent clusters that can be bridged to the high-reward manifold via specific translational vectors.}
    \label{fig:tsne}
    \vspace{-0.3cm}
\end{figure}

To elucidate the mechanistic basis of STIR, we visualize the geometry of the latent state embeddings using t-SNE~\cite{maaten2008visualizing} projections. Figure~\ref{fig:tsne} maps the spatial relationship between the failure states (negative samples) and their corresponding successful counterparts (positive samples) across the MATH500 and OBQA benchmarks. We observe two distinct structural phenomena. First, the embeddings of reasoning failures are not uniformly distributed but organize into coherent clusters, indicating that different error modes occupy distinct regions in the latent manifold.
This clustering suggests that reasoning failures collapse into low-dimensional attractor basins, providing stable targets for intervention.
Second, the spatial transition from negative to positive samples exhibits consistent directionality within these local neighborhoods.
Such geometric alignment implies that the validity of a reasoning step is encoded as a linear feature in the residual stream.
This geometric structure confirms that our steering impulses function as precise translational vectors that bridge the gap between specific error modes and the high-reward subspace, validating the effectiveness of our intrinsic action induction.

\subsection{Sensitivity Analysis}
\label{sec:sensitivity}

\begin{figure*}[t]
    \centering
    \begin{subfigure}{0.24\textwidth}
        \includegraphics[width=\linewidth]{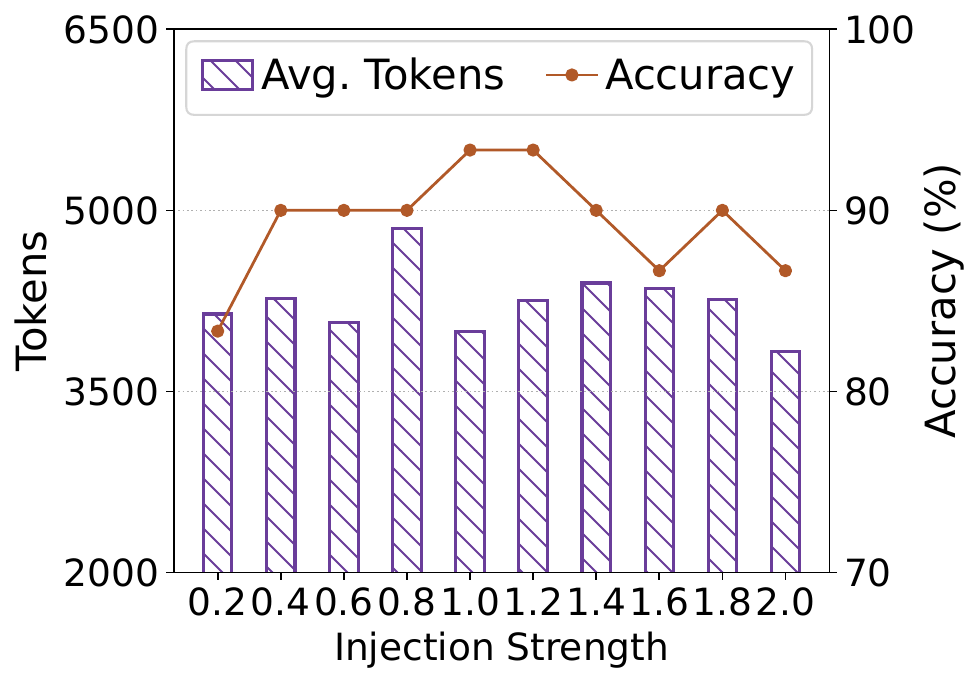}
        \caption{$k_{scale}$ on AMC23}
    \end{subfigure}
    \begin{subfigure}{0.24\textwidth}
        \includegraphics[width=\linewidth]{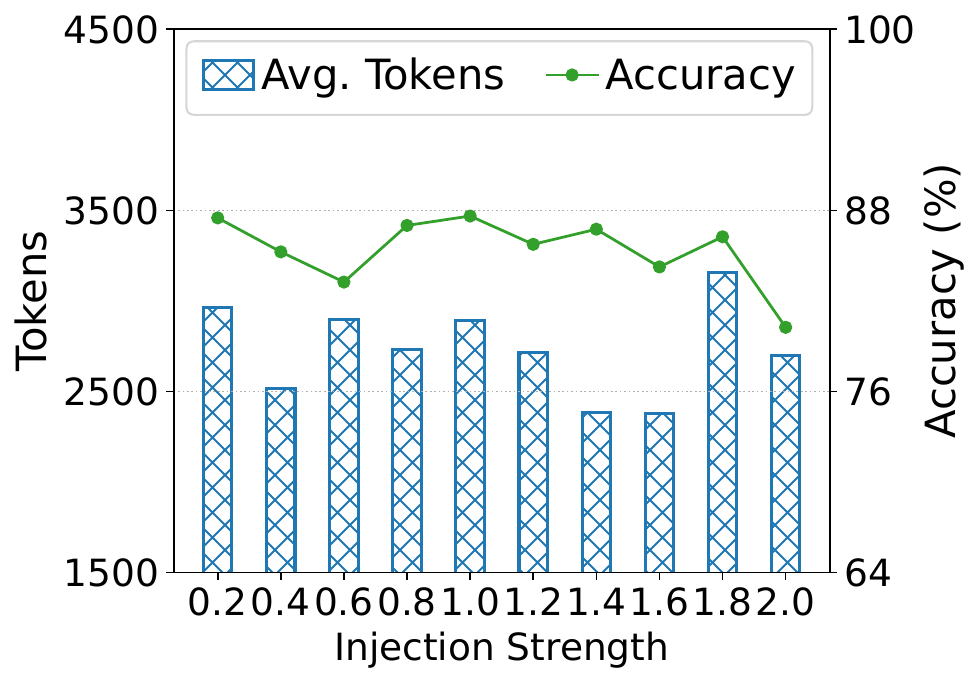}
        \caption{$k_{scale}$ on MATH500}
    \end{subfigure}
    \begin{subfigure}{0.24\textwidth}
        \includegraphics[width=\linewidth]{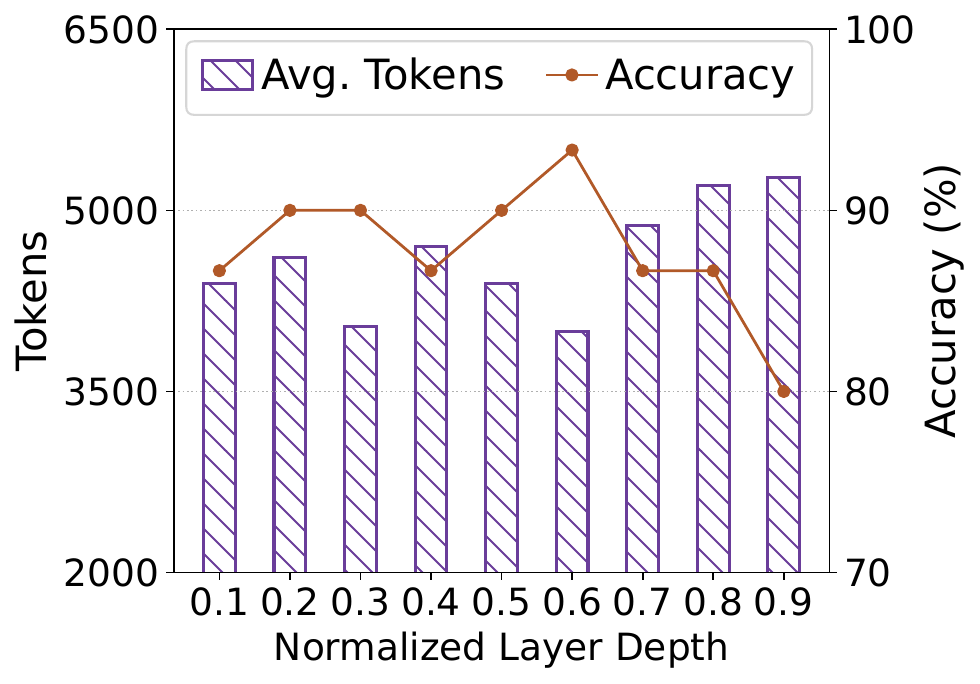}
        \caption{Layer on AMC23}
    \end{subfigure}
    \begin{subfigure}{0.24\textwidth}
        \includegraphics[width=\linewidth]{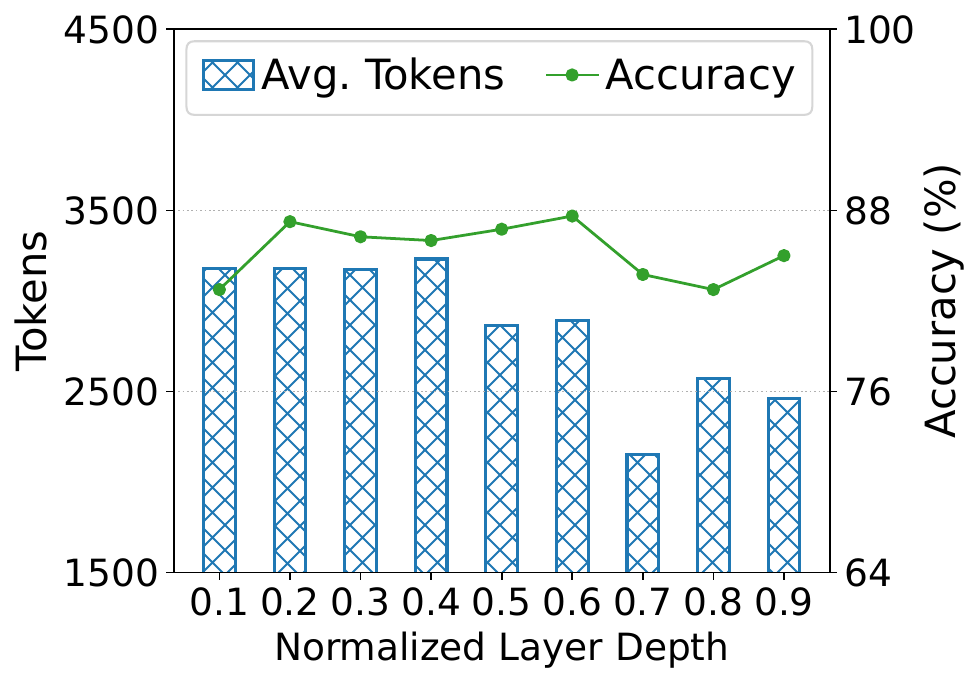}
        \caption{Layer on MATH500}
    \end{subfigure}
    \caption{Sensitivity analysis of key control hyperparameters on AMC23 and MATH500. Subplots (a) and (b) demonstrate the inverted-U relationship between injection strength $k_{\text{scale}}$ and reasoning accuracy. Subplots (c) and (d) illustrate the impact of normalized layer depth and reveal that steering interventions are most effective within the intermediate transformer blocks.}
    \label{fig:sensitivity}
\end{figure*}

To verify the operational stability of STIR, we analyze the system's response to variations in two governing hyperparameters: the injection strength $k_{scale}$ and the normalized layer depth. Figure~\ref{fig:sensitivity} visualizes these dynamics across the AMC23 and MATH500 benchmarks. Regarding injection strength, we observe a consistent inverted-U performance trend. Weak perturbations ($k_{scale} < 0.5$) often fail to overcome the inertia of the default erroneous trajectory and result in marginal gains. Conversely, excessive magnitudes ($k_{scale} > 1.5$) tend to push latent states off the valid semantic manifold and precipitate a collapse in linguistic coherence. The optimal control window lies reliably within $k_{scale} \in [0.75, 1.25]$ and confirms that our value-modulated mechanism effectively normalizes the steering impulse across different contexts. In terms of layer selectivity, performance consistently peaks when interventions target the intermediate transformer blocks (relative depth $0.4$--$0.6$). This aligns with the mechanistic intuition that early layers process low-level syntax while late layers focus on token prediction, locating the critical logical transitions susceptible to steering in the middle of the network depth.

\subsection{Ablation Study}
\label{sec:ablation}

To validate the architectural decisions in STIR, we conduct a comprehensive component-wise ablation study on the AMC23 and MATH500 benchmarks. Table~\ref{tab:ablation} summarizes the impact of selectively removing key modules. Regarding offline memory construction, replacing our contrastive intrinsic action induction with random sampling precipitates a significant accuracy drop, confirming that valid steering impulses must be distilled from the differential dynamics between successful and failed rollouts. Furthermore, reverting to naive top-k selection degrades performance, particularly on AMC23, as it saturates the memory with redundant failure modes and fails to span the geometric manifold of reasoning errors. In terms of the online control policy, removing lookahead probing harms accuracy, indicating that static retrieval similarity alone is insufficient to guarantee intervention utility. Notably, disabling the anchor gating mechanism leads to over-intervention on optimal trajectories, resulting in a substantial increase in computational cost and disrupted reasoning coherence. Collectively, these findings confirm that the synergy between diversity-aware memory and evidence-based gating is critical for establishing a superior accuracy-efficiency Pareto frontier.

\begin{table}[t]
    \centering
    \caption{Ablation study on AMC23 and MATH500 benchmarks. We report the impact of removing key modules from the offline memory construction and online control policy. The results demonstrate that each component is essential for maintaining the optimal balance between reasoning accuracy and computational efficiency.}
    \label{tab:ablation}
    
    \setlength{\tabcolsep}{2.5pt}
    \renewcommand{\arraystretch}{1.2}
    
    \resizebox{\columnwidth}{!}{
    \begin{tabular}{l cc cc}
        \toprule
        \multirow{2}{*}{\textbf{Configuration}} & \multicolumn{2}{c}{\textbf{AMC 23}} & \multicolumn{2}{c}{\textbf{MATH 500}} \\
        \cmidrule(lr){2-3} \cmidrule(lr){4-5}
         & \textbf{Acc.} & \textbf{Avg. Cost} & \textbf{Acc.} & \textbf{Avg. Cost} \\
        \midrule
        
        \rowcolor{gray!10} 
        \textbf{STIR (Full)} & \textbf{93.3} & \textbf{4000} & \textbf{88.0} & \textbf{2639} \\
        \midrule
        
        \multicolumn{5}{l}{\textit{Offline Memory Construction}} \\
        \hspace{1em} w/o Contrastive Mining & 
            83.3 \scriptsize{(\textcolor{rcolor}{-10.0})} & 5041 \scriptsize{(\textcolor{rcolor}{+1041})} & 
            83.0 \scriptsize{(\textcolor{rcolor}{-5.0})} & 2904 \scriptsize{(\textcolor{rcolor}{+265})} \\
            
        \hspace{1em} w/o Diversity (Top-k) & 
            76.7 \scriptsize{(\textcolor{rcolor}{-16.6})} & 6048 \scriptsize{(\textcolor{rcolor}{+2048})} & 
            87.3 \scriptsize{(\textcolor{rcolor}{-0.7})} & 2543 \scriptsize{(\textcolor{bcolor}{-96})} \\
        \midrule
        
        \multicolumn{5}{l}{\textit{Online Control Policy}} \\
        \hspace{1em} w/o Lookahead Probing & 
            83.3 \scriptsize{(\textcolor{rcolor}{-10.0})} & 5180 \scriptsize{(\textcolor{rcolor}{+1180})} & 
            81.0 \scriptsize{(\textcolor{rcolor}{-7.0})} & 2096 \scriptsize{(\textcolor{bcolor}{-543})} \\
            
        \hspace{1em} w/o Anchor Gating & 
            86.7 \scriptsize{(\textcolor{rcolor}{-6.6})} & 5203 \scriptsize{(\textcolor{rcolor}{+1203})} & 
            85.8 \scriptsize{(\textcolor{rcolor}{-2.2})} & 2855 \scriptsize{(\textcolor{rcolor}{+216})} \\
            
        \hspace{1em} w/o Adaptive Injection & 
            93.3 \scriptsize{(\textcolor{rcolor}{-0.0})} & 4488 \scriptsize{(\textcolor{rcolor}{+488})} & 
            80.3 \scriptsize{(\textcolor{rcolor}{-7.7})} & 3405 \scriptsize{(\textcolor{rcolor}{+766})} \\
        \bottomrule
    \end{tabular}
    }
\end{table}

\subsection{Cross-Task Transferability of Latent Tools}
\label{sec:generalization}

\begin{figure*}
    \centering
    \includegraphics[width=0.95\linewidth]{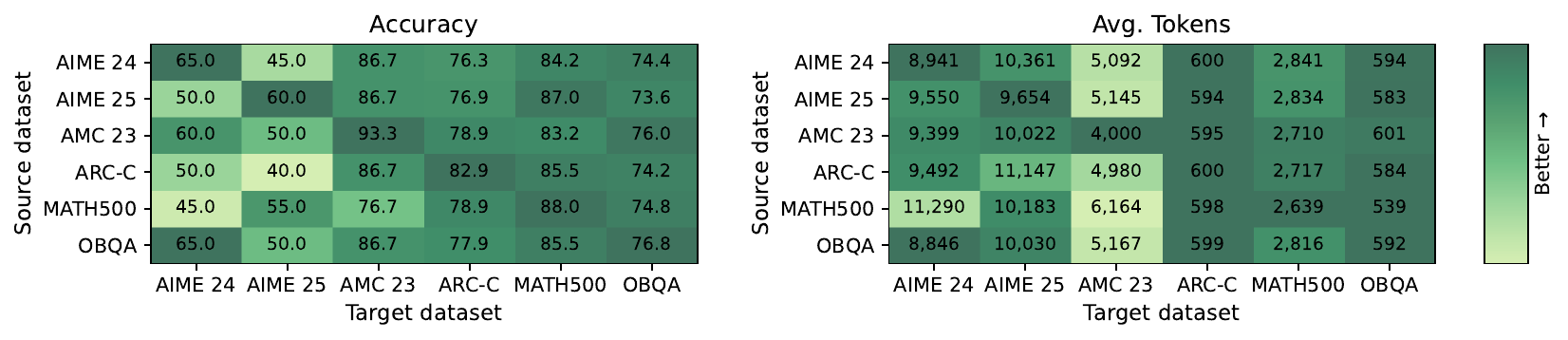}
    \caption{Cross-task generalization analysis. The heatmaps depict the transfer performance when a tool library distilled from a source dataset (y-axis) is applied to a target dataset (x-axis). The left panel reports reasoning accuracy, while the right panel shows the average token count. The strong performance in off-diagonal entries validates that STIR captures transferable latent tools that generalize across distinct tasks and domains, demonstrating robustness against specific problem distributions.}
    \label{fig:generalization}
\end{figure*}

\begin{table}[h]
    \centering
    \caption{Computational overhead analysis across model scales. Avg. tokens denotes the mean cumulative count of additional tokens required for the retrieve-preview-commit cycle, while ratio represents this overhead as a percentage of the total inference cost.}
    \label{tab:overhead}
    \setlength{\tabcolsep}{2.5pt} 
    \renewcommand{\arraystretch}{1.1} 
    \resizebox{\columnwidth}{!}{
    \begin{tabular}{l cc cc}
        \toprule
        \multirow{2}{*}{\textbf{Datasets}} & \multicolumn{2}{c}{\textbf{DeepSeek-R1-1.5B}} & \multicolumn{2}{c}{\textbf{DeepSeek-R1-7B}} \\
        \cmidrule(lr){2-3} \cmidrule(lr){4-5}
        & \textbf{Avg.(Total) Tokens} & \textbf{Ratio} & \textbf{Avg.(Total) Tokens} & \textbf{Ratio} \\
        \midrule
        AIME 24& 138.80 (11930) & 1.16\%& 130.51 (8941) & 1.46\%
\\
        AIME 25& 134.80 (11383) & 1.18\%& 138.54 (9653) & 1.44\%
\\
        AMC 23& 140.00 (6645) & 2.11\%& 56.00 (4000) & 1.40\%
\\
        MATH 500& 128.79 (4297) & 3.00\%& 132.84 (2638) & 5.03\%
\\
        \bottomrule
    \end{tabular}
    }
\end{table}

To investigate whether the distilled latent tools capture universal reasoning patterns, we conduct a comprehensive cross-domain transfer study. We construct the tool library using a single source dataset and evaluate its zero-shot performance across all target benchmarks. Figure~\ref{fig:generalization} illustrates the resulting transfer matrices for accuracy and computational cost. We observe robust off-diagonal performance, particularly within the mathematical domain, where tools distilled from rigorous tasks like AIME effectively steer trajectories in AMC 23 and MATH500. Notably, even cross-domain transfer between mathematical reasoning and general QA yields positive gains. This finding suggests that STIR identifies task-agnostic cognitive functions, including hypothesis verification and logical decomposition, which are fundamental to general reasoning. Furthermore, the reduction in token consumption remains consistent across settings. This consistency confirms that the structural efficiency of our control policy stems from the intrinsic properties of the tools, independent of dataset-specific correlations.

\subsection{Computational Overhead Analysis}
\label{sec:overhead}
We rigorously quantify the computational cost of STIR by tracking the cumulative token consumption incurred by the retrieve-preview-commit control cycle. As detailed in Table~\ref{tab:overhead}, the framework introduces a mean overhead of approximately 130 to 140 tokens per query across the evaluated benchmarks. Crucially, relative to the extensive chain-of-thought trajectories which typically span thousands of tokens, this translates to a marginal relative cost of 1.1\% to 5.0\%. This structural efficiency is architecturally underpinned by our batched lookahead implementation, which leverages shared KV-caching to validate multiple candidate tools in parallel. Instead of re-processing the context for each potential steering direction, we reuse the pre-computed key-value states of the prefix to execute counterfactual probing within a single batched forward pass. This design effectively eliminates redundant re-computation, ensuring that high-precision trajectory steering is achieved with a minimal and scalable computational footprint.

\section{Conclusion}

In this work, we present STIR, a unified framework that bridges the gap between static activation steering and the non-stationary dynamics of complex LLM reasoning. By orchestrating differential intrinsic action induction, sparse control basis construction, and value-modulated trajectory intervention, our approach distills transient internal reasoning successes into a retrievable episodic tool memory. This mechanism effectively internalizes the benefits of explicit search and self-correction directly within the residual stream, fundamentally decoupling reasoning depth from sequence length. Extensive experiments demonstrate that STIR establishes a new Pareto frontier in accuracy-efficiency trade-offs, significantly outperforming standard decoding baselines.
Collectively, these findings enable the compaction of explicit system 2 reasoning processes into implicit system 1 latent dynamics, thereby laying the foundation for a new paradigm of efficient and high-fidelity cognitive modeling that effectively bridges the gap between the rigor of explicit search and the efficiency of neural intuition.

\bibliography{main}

@inproceedings{kwon2023efficient,
  title={Efficient Memory Management for Large Language Model Serving with PagedAttention},
  author={Woosuk Kwon and Zhuohan Li and Siyuan Zhuang and Ying Sheng and Lianmin Zheng and Cody Hao Yu and Joseph E. Gonzalez and Hao Zhang and Ion Stoica},
  booktitle={Proceedings of the ACM SIGOPS 29th Symposium on Operating Systems Principles},
  year={2023}
}

@article{xu2025easysteer,
  title={EasySteer: A Unified Framework for High-Performance and Extensible LLM Steering},
  author={Xu, Haolei and Mei, Xinyu and Yan, Yuchen and Zhou, Rui and Zhang, Wenqi and Lu, Weiming and Zhuang, Yueting and Shen, Yongliang},
  journal={arXiv preprint arXiv:2509.25175},
  year={2025}
}

@inproceedings{mihaylov2018can,
  title={Can a Suit of Armor Conduct Electricity? A New Dataset for Open Book Question Answering},
  author={Mihaylov, Todor and Clark, Peter and Khot, Tushar and Sabharwal, Ashish},
  booktitle={Proceedings of the 2018 Conference on Empirical Methods in Natural Language Processing},
  pages={2381--2391},
  year={2018}
}

@article{clark2018think,
  title={Think you have solved question answering? try arc, the ai2 reasoning challenge},
  author={Clark, Peter and Cowhey, Isaac and Etzioni, Oren and Khot, Tushar and Sabharwal, Ashish and Schoenick, Carissa and Tafjord, Oyvind},
  journal={arXiv preprint arXiv:1803.05457},
  year={2018}
}

@article{hendrycksmath2021,
  title={Measuring Mathematical Problem Solving With the MATH Dataset},
  author={Dan Hendrycks and Collin Burns and Saurav Kadavath and Akul Arora and Steven Basart and Eric Tang and Dawn Song and Jacob Steinhardt},
  journal={Advances in Neural Information Processing Systems},
  year={2021}
}

@article{guo2025deepseek,
  title={Deepseek-r1: Incentivizing reasoning capability in llms via reinforcement learning},
  author={Guo, Daya and Yang, Dejian and Zhang, Haowei and Song, Junxiao and Zhang, Ruoyu and Xu, Runxin and Zhu, Qihao and Ma, Shirong and Wang, Peiyi and Bi, Xiao and others},
  journal={arXiv preprint arXiv:2501.12948},
  year={2025}
}

@article{yang2025qwen3,
  title={Qwen3 technical report},
  author={Yang, An and Li, Anfeng and Yang, Baosong and Zhang, Beichen and Hui, Binyuan and Zheng, Bo and Yu, Bowen and Gao, Chang and Huang, Chengen and Lv, Chenxu and others},
  journal={arXiv preprint arXiv:2505.09388},
  year={2025}
}

@article{zhou2024self,
  title={Self-discover: Large language models self-compose reasoning structures},
  author={Zhou, Pei and Pujara, Jay and Ren, Xiang and Chen, Xinyun and Cheng, Heng-Tze and Le, Quoc V and Chi, Ed and Zhou, Denny and Mishra, Swaroop and Zheng, Huaixiu Steven},
  journal={Advances in Neural Information Processing Systems},
  volume={37},
  pages={126032--126058},
  year={2024}
}

@article{wang2022self,
  title={Self-consistency improves chain of thought reasoning in language models},
  author={Wang, Xuezhi and Wei, Jason and Schuurmans, Dale and Le, Quoc and Chi, Ed and Narang, Sharan and Chowdhery, Aakanksha and Zhou, Denny},
  journal={arXiv preprint arXiv:2203.11171},
  year={2022}
}

@article{chen2024more,
  title={Are more llm calls all you need? towards the scaling properties of compound ai systems},
  author={Chen, Lingjiao and Davis, Jared and Hanin, Boris and Bailis, Peter and Stoica, Ion and Zaharia, Matei and Zou, James},
  journal={Advances in Neural Information Processing Systems},
  volume={37},
  pages={45767--45790},
  year={2024}
}

@article{yang2025dynamic,
  title={Dynamic Early Exit in Reasoning Models},
  author={Yang, Chenxu and Si, Qingyi and Duan, Yongjie and Zhu, Zheliang and Zhu, Chenyu and Lin, Zheng and Cao, Li and Wang, Weiping},
  journal={arXiv preprint arXiv:2504.15895},
  year={2025}
}

@article{chen2025seal,
  title={Seal: Steerable reasoning calibration of large language models for free},
  author={Chen, Runjin and Zhang, Zhenyu and Hong, Junyuan and Kundu, Souvik and Wang, Zhangyang},
  journal={arXiv preprint arXiv:2504.07986},
  year={2025}
}

@article{kulesza2012determinantal,
  title={Determinantal point processes for machine learning},
  author={Kulesza, Alex and Taskar, Ben and others},
  journal={Foundations and Trends{\textregistered} in Machine Learning},
  volume={5},
  number={2--3},
  pages={123--286},
  year={2012}
}

@article{gillenwater2012near,
  title={Near-optimal map inference for determinantal point processes},
  author={Gillenwater, Jennifer and Kulesza, Alex and Taskar, Ben},
  journal={Advances in Neural Information Processing Systems},
  volume={25},
  year={2012}
}

@article{maaten2008visualizing,
  title={Visualizing data using t-SNE},
  author={Maaten, Laurens van der and Hinton, Geoffrey},
  journal={Journal of machine learning research},
  volume={9},
  number={Nov},
  pages={2579--2605},
  year={2008}
}

@article{wei2022chain,
  title={Chain-of-thought prompting elicits reasoning in large language models},
  author={Wei, Jason and Wang, Xuezhi and Schuurmans, Dale and Bosma, Maarten and Xia, Fei and Chi, Ed and Le, Quoc V and Zhou, Denny and others},
  journal={Advances in neural information processing systems},
  volume={35},
  pages={24824--24837},
  year={2022}
}

@article{kojima2022large,
  title={Large language models are zero-shot reasoners},
  author={Kojima, Takeshi and Gu, Shixiang Shane and Reid, Machel and Matsuo, Yutaka and Iwasawa, Yusuke},
  journal={Advances in neural information processing systems},
  volume={35},
  pages={22199--22213},
  year={2022}
}

@article{yao2023tree,
  title={Tree of thoughts: Deliberate problem solving with large language models},
  author={Yao, Shunyu and Yu, Dian and Zhao, Jeffrey and Shafran, Izhak and Griffiths, Tom and Cao, Yuan and Narasimhan, Karthik},
  journal={Advances in neural information processing systems},
  volume={36},
  pages={11809--11822},
  year={2023}
}

@inproceedings{besta2024graph,
  title={Graph of thoughts: Solving elaborate problems with large language models},
  author={Besta, Maciej and Blach, Nils and Kubicek, Ales and Gerstenberger, Robert and Podstawski, Michal and Gianinazzi, Lukas and Gajda, Joanna and Lehmann, Tomasz and Niewiadomski, Hubert and Nyczyk, Piotr and others},
  booktitle={Proceedings of the AAAI conference on artificial intelligence},
  volume={38},
  number={16},
  pages={17682--17690},
  year={2024}
}

@article{snell2024scaling,
  title={Scaling llm test-time compute optimally can be more effective than scaling model parameters},
  author={Snell, Charlie and Lee, Jaehoon and Xu, Kelvin and Kumar, Aviral},
  journal={arXiv preprint arXiv:2408.03314},
  year={2024}
}

@article{zelikman2022star,
  title={Star: Bootstrapping reasoning with reasoning},
  author={Zelikman, Eric and Wu, Yuhuai and Mu, Jesse and Goodman, Noah},
  journal={Advances in Neural Information Processing Systems},
  volume={35},
  pages={15476--15488},
  year={2022}
}

@article{lin2025implicit,
  title={Implicit reasoning in transformers is reasoning through shortcuts},
  author={Lin, Tianhe and Xie, Jian and Yuan, Siyu and Yang, Deqing},
  journal={arXiv preprint arXiv:2503.07604},
  year={2025}
}

@article{hao2024training,
  title={Training large language models to reason in a continuous latent space},
  author={Hao, Shibo and Sukhbaatar, Sainbayar and Su, DiJia and Li, Xian and Hu, Zhiting and Weston, Jason and Tian, Yuandong},
  journal={arXiv preprint arXiv:2412.06769},
  year={2024}
}

@article{skean2025layer,
  title={Layer by layer: Uncovering hidden representations in language models},
  author={Skean, Oscar and Arefin, Md Rifat and Zhao, Dan and Patel, Niket and Naghiyev, Jalal and LeCun, Yann and Shwartz-Ziv, Ravid},
  journal={arXiv preprint arXiv:2502.02013},
  year={2025}
}

@article{zelikman2024quiet,
  title={Quiet-star: Language models can teach themselves to think before speaking},
  author={Zelikman, Eric and Harik, Georges and Shao, Yijia and Jayasiri, Varuna and Haber, Nick and Goodman, Noah D},
  journal={arXiv preprint arXiv:2403.09629},
  year={2024}
}

@article{kim2025learning,
  title={Learning to Insert [PAUSE] Tokens for Better Reasoning},
  author={Kim, Eunki and Kim, Sangryul and Thorne, James},
  journal={arXiv preprint arXiv:2506.03616},
  year={2025}
}

@article{london2025pause,
  title={Pause Tokens Strictly Increase the Expressivity of Constant-Depth Transformers},
  author={London, Charles and Kanade, Varun},
  journal={arXiv preprint arXiv:2505.21024},
  year={2025}
}

@article{zhu2025survey,
  title={A survey on latent reasoning},
  author={Zhu, Rui-Jie and Peng, Tianhao and Cheng, Tianhao and Qu, Xingwei and Huang, Jinfa and Zhu, Dawei and Wang, Hao and Xue, Kaiwen and Zhang, Xuanliang and Shan, Yong and others},
  journal={arXiv preprint arXiv:2507.06203},
  year={2025}
}

@article{shen2025codi,
  title={Codi: Compressing chain-of-thought into continuous space via self-distillation},
  author={Shen, Zhenyi and Yan, Hanqi and Zhang, Linhai and Hu, Zhanghao and Du, Yali and He, Yulan},
  journal={arXiv preprint arXiv:2502.21074},
  year={2025}
}

@article{turner2023steering,
  title={Steering language models with activation engineering},
  author={Turner, Alexander Matt and Thiergart, Lisa and Leech, Gavin and Udell, David and Vazquez, Juan J and Mini, Ulisse and MacDiarmid, Monte},
  journal={arXiv preprint arXiv:2308.10248},
  year={2023}
}

@inproceedings{rimsky2024steering,
  title={Steering llama 2 via contrastive activation addition},
  author={Rimsky, Nina and Gabrieli, Nick and Schulz, Julian and Tong, Meg and Hubinger, Evan and Turner, Alexander},
  booktitle={Proceedings of the 62nd Annual Meeting of the Association for Computational Linguistics (Volume 1: Long Papers)},
  pages={15504--15522},
  year={2024}
}

@article{stolfo2024improving,
  title={Improving instruction-following in language models through activation steering},
  author={Stolfo, Alessandro and Balachandran, Vidhisha and Yousefi, Safoora and Horvitz, Eric and Nushi, Besmira},
  journal={arXiv preprint arXiv:2410.12877},
  year={2024}
}

@article{azizi2025activation,
  title={Activation steering for chain-of-thought compression},
  author={Azizi, Seyedarmin and Potraghloo, Erfan Baghaei and Pedram, Massoud},
  journal={arXiv preprint arXiv:2507.04742},
  year={2025}
}

@article{huang2025mitigating,
  title={Mitigating Overthinking in Large Reasoning Models via Manifold Steering},
  author={Huang, Yao and Chen, Huanran and Ruan, Shouwei and Zhang, Yichi and Wei, Xingxing and Dong, Yinpeng},
  journal={arXiv preprint arXiv:2505.22411},
  year={2025}
}

@article{todd2023function,
  title={Function vectors in large language models},
  author={Todd, Eric and Li, Millicent L and Sharma, Arnab Sen and Mueller, Aaron and Wallace, Byron C and Bau, David},
  journal={arXiv preprint arXiv:2310.15213},
  year={2023}
}

@article{zhang2025survey,
  title={A Survey on Test-Time Scaling in Large Language Models: What, How, Where, and How Well?},
  author={Zhang, Qiyuan and Lyu, Fuyuan and Sun, Zexu and Wang, Lei and Zhang, Weixu and Hua, Wenyue and Wu, Haolun and Guo, Zhihan and Wang, Yufei and Muennighoff, Niklas and others},
  journal={arXiv preprint arXiv:2503.24235},
  year={2025}
}

@article{zhang2024rest,
  title={Rest-mcts*: Llm self-training via process reward guided tree search},
  author={Zhang, Dan and Zhoubian, Sining and Hu, Ziniu and Yue, Yisong and Dong, Yuxiao and Tang, Jie},
  journal={Advances in Neural Information Processing Systems},
  volume={37},
  pages={64735--64772},
  year={2024}
}

@article{feng2025optimal,
  title={Optimal Self-Consistency for Efficient Reasoning with Large Language Models},
  author={Feng, Austin and Alonso, Marius and Odonnat, Ambroise},
  journal={arXiv preprint arXiv:2511.12309},
  year={2025}
}

@article{chen2025reasoning,
  title={Reasoning Beyond Language: A Comprehensive Survey on Latent Chain-of-Thought Reasoning},
  author={Chen, Xinghao and Zhao, Anhao and Xia, Heming and Lu, Xuan and Wang, Hanlin and Chen, Yanjun and Zhang, Wei and Wang, Jian and Li, Wenjie and Shen, Xiaoyu},
  journal={arXiv preprint arXiv:2505.16782},
  year={2025}
}

@article{zhou2022least,
  title={Least-to-most prompting enables complex reasoning in large language models},
  author={Zhou, Denny and Sch{\"a}rli, Nathanael and Hou, Le and Wei, Jason and Scales, Nathan and Wang, Xuezhi and Schuurmans, Dale and Cui, Claire and Bousquet, Olivier and Le, Quoc and others},
  journal={arXiv preprint arXiv:2205.10625},
  year={2022}
}

@article{zeng2025revisiting,
  title={Revisiting the Test-Time Scaling of o1-like Models: Do they Truly Possess Test-Time Scaling Capabilities?},
  author={Zeng, Zhiyuan and Cheng, Qinyuan and Yin, Zhangyue and Zhou, Yunhua and Qiu, Xipeng},
  journal={arXiv preprint arXiv:2502.12215},
  year={2025}
}

@article{sui2025stop,
  title={Stop overthinking: A survey on efficient reasoning for large language models},
  author={Sui, Yang and Chuang, Yu-Neng and Wang, Guanchu and Zhang, Jiamu and Zhang, Tianyi and Yuan, Jiayi and Liu, Hongyi and Wen, Andrew and Zhong, Shaochen and Zou, Na and others},
  journal={arXiv preprint arXiv:2503.16419},
  year={2025}
}

@article{zhu2025scaling,
  title={Scaling latent reasoning via looped language models},
  author={Zhu, Rui-Jie and Wang, Zixuan and Hua, Kai and Zhang, Tianyu and Li, Ziniu and Que, Haoran and Wei, Boyi and Wen, Zixin and Yin, Fan and Xing, He and others},
  journal={arXiv preprint arXiv:2510.25741},
  year={2025}
}

@article{zhang2025latent,
  title={Do latent tokens think? a causal and adversarial analysis of chain-of-continuous-thought},
  author={Zhang, Yuyi and Tang, Boyu and Ju, Tianjie and Duan, Sufeng and Liu, Gongshen},
  journal={arXiv preprint arXiv:2512.21711},
  year={2025}
}

@article{su2025token,
  title={Token assorted: Mixing latent and text tokens for improved language model reasoning},
  author={Su, DiJia and Zhu, Hanlin and Xu, Yingchen and Jiao, Jiantao and Tian, Yuandong and Zheng, Qinqing},
  journal={arXiv preprint arXiv:2502.03275},
  year={2025}
}

@article{turner2024activation,
  title={Activation addition: Steering language models without optimization},
  author={Turner, Alexander Matt and Thiergart, Lisa and Leech, Gavin and Udell, David and Mini, Ulisse and MacDiarmid, Monte},
  year={2024}
}

@article{bartoszcze2025representation,
  title={Representation Engineering for Large-Language Models: Survey and Research Challenges},
  author={Bartoszcze, Lukasz and Munshi, Sarthak and Sukidi, Bryan and Yen, Jennifer and Yang, Zejia and Williams-King, David and Le, Linh and Asuzu, Kosi and Maple, Carsten},
  journal={arXiv preprint arXiv:2502.17601},
  year={2025}
}

@article{chen2025persona,
  title={Persona vectors: Monitoring and controlling character traits in language models},
  author={Chen, Runjin and Arditi, Andy and Sleight, Henry and Evans, Owain and Lindsey, Jack},
  journal={arXiv preprint arXiv:2507.21509},
  year={2025}
}

@article{park2023linear,
  title={The linear representation hypothesis and the geometry of large language models},
  author={Park, Kiho and Choe, Yo Joong and Veitch, Victor},
  journal={arXiv preprint arXiv:2311.03658},
  year={2023}
}

@article{ansuini2019intrinsic,
  title={Intrinsic dimension of data representations in deep neural networks},
  author={Ansuini, Alessio and Laio, Alessandro and Macke, Jakob H and Zoccolan, Davide},
  journal={Advances in Neural Information Processing Systems},
  volume={32},
  year={2019}
}

@article{loaiza2024deep,
  title={Deep generative models through the lens of the manifold hypothesis: A survey and new connections},
  author={Loaiza-Ganem, Gabriel and Ross, Brendan Leigh and Hosseinzadeh, Rasa and Caterini, Anthony L and Cresswell, Jesse C},
  journal={arXiv preprint arXiv:2404.02954},
  year={2024}
}

@misc{aime,
    title = {AIME Problems and Solutions},
    author={{MAA Committees}},
    howpublished = {\url{https://artofproblemsolving.com/wiki/index.php/AIME_Problems_and_Solutions}}
}

@misc{amc2023,
  title={American Mathematics Competition 2023 Dataset},
  author={{math-ai}},
  howpublished={\url{https://huggingface.co/datasets/math-ai/amc23}}
}
\bibliographystyle{icml2026}

\appendix

\onecolumn

\section{Implementation Details}
\label{app:implementation_details}

This appendix presents the technical specifications of STIR, detailing the mathematical notations and hyperparameter settings used in our experiments. The implementation description is structured around the three core components: differential intrinsic action induction, sparse control basis construction, and value-modulated trajectory intervention. These details supplement the methodology in Section~\ref{sec:method} and provide the necessary parameters to reproduce the results.

\subsection{Notation Summary}
\label{app:notation}

Table~\ref{tab:notation} summarizes the mathematical symbols used to describe the interaction between the language model and the episodic memory. These notations represent the system's components, ranging from the residual stream dynamics to the geometric structure of the control basis. The table specifically distinguishes between the two functional memory units derived during training: correction entries, which contain active steering impulses to rectify reasoning errors, and anchor entries, which serve as reference points for optimal states to support the gating mechanism.

\begin{table}[h]
\centering
\caption{Mathematical notations and their descriptions used in the STIR framework.}
\label{tab:notation}
\begin{tabular}{c|p{0.7\linewidth}}
\toprule
\textbf{Symbol} & \textbf{Description} \\
\midrule
$\mathcal{M}_{\theta}$ & Target transformer-based Large Language Model. \\
$h_{t}^{(l)}$ & Latent hidden state at timestep $t$ and layer $l$. \\
$\pi=(k,v,q)$ & Steering tool unit comprising state key, steering impulses, and quality score. \\
$k \in \mathbb{R}^{d}$ & State key (context anchor) used for tool retrieval. \\
$v \in \mathbb{R}^{d}$ & Steering impulse (directional vector) for trajectory correction. \\
$q(\pi)$ & Historical quality score derived from reward gain during offline discovery. \\
$\mathcal{M}$ & Sparse control basis representing the episodic memory repository. \\
$\Omega$ & Raw candidate set of discovered tools before sparse selection. \\
$\mathcal{C}_{t}$ & Locally retrieved candidate set of steering tools at timestep $t$. \\
$\hat{A}(\pi)$ & Similarity-weighted quality score representing the retrieval prior. \\
$\mathcal{G}(\pi)$ & Dynamic gain estimated via counterfactual lookahead probing. \\
$S(\pi)$ & Unified utility score for final tool selection and gating. \\
$\tau_{null}$ & Hard abstention threshold (null action threshold). \\
$k_{scale}$ & Scaling factor for adaptive injection strength. \\
$\beta, \rho$ & Weighting coefficients for prior score and probing gain, respectively. \\
$\lambda$ & Diversity penalty coefficient in the sparse basis construction objective. \\
$T_{probe}$ & Horizon length (tokens) for lookahead previewing. \\
$B$ & Total memory budget (maximum entries in the library). \\
\bottomrule
\end{tabular}
\end{table}

\subsection{Hyperparameter Settings and Configuration}
\label{subsec:hyperparameters}

Table~\ref{tab:hyperparams} details the hyperparameter configurations employed in the STIR framework. For offline memory construction, we set the stochastic rollout budget $K=64$ and the length penalty coefficient $\eta=0.5$. These values are selected to prioritize the discovery of concise reasoning paths while ensuring sufficient coverage of potential failure modes. The sparse control basis is constructed with a maximum memory budget $B=500$ and a diversity penalty $\lambda=0.5$, which enforces geometric orthogonality among the selected steering primitives. In the online control phase, the retrieval pool size is set to $k=16$ and the lookahead horizon to $T_{probe}=8$, a configuration calibrated to optimize the trade-off between the precision of the dynamic gain estimation and the inference latency.

\begin{table}[h]
\centering
\caption{Hyperparameter configurations employed in the STIR framework.}
\label{tab:hyperparams}
\begin{tabular}{c|c|c}
\toprule
\textbf{Phase} & \textbf{Hyperparameter} & \textbf{Value} \\
\midrule
\multirow{5}{*}{\shortstack[l]{Differential Intrinsic\\Action Induction}} 
 & Rollout count per example ($K$) & 8 \\
 & Positive pool size ($k_{pos}$) & 2 \\
 & Negative pool size ($k_{neg}$) & 3 \\
 & Length penalty coefficient ($\eta$) & 0.01 \\
 & Sampling temperature & 0.7 \\
\midrule
\multirow{3}{*}{\shortstack[l]{Sparse Control\\Basis Construction}} 
 & Total library size ($B$) & 256 \\
 & Diversity weight ($\lambda$) & 0.5 \\
 & DPP stability term ($\epsilon$) & $10^{-4}$ \\
\midrule
\multirow{7}{*}{\shortstack[l]{Value-Modulated\\Trajectory Intervention}} 
 & Retrieval top-$k$ & 8 \\
 & Candidate pool size ($L$) & 4 \\
 & Lookahead horizon ($T_{probe}$) & 4 \\
 & Prior weight ($\beta$) & 2.0 \\
 & Probing weight ($\rho$) & 0.1 \\
 & Normalized layer depth & 0.6 \\
 & Strength scale ($k_{scale}$) & 1.0 \\
\bottomrule
\end{tabular}
\end{table}

\begin{algorithm}[htbp]
   \caption{STIR Offline: Differential Action Induction and Sparse Basis Construction}
   \label{alg:offline}
\begin{algorithmic}[1]
   \REQUIRE Training set $\mathcal{D}$, model $\mathcal{M}_{\theta}$, checkpoints $M$, memory budget $B$
   \ENSURE Sparse control basis $\mathcal{M}$
   \STATE // Stage I: Contrastive rollout discovery
   \FOR{each example $(x,y) \in \mathcal{D}$}
       \STATE Sample $K$ stochastic rollouts $\{Y_1, ..., Y_K\}$ from $\mathcal{M}_{\theta}$
       \STATE Compute rewards $\{R_1, ..., R_K\}$ using Eq. (3)
       \FOR{each checkpoint $m=1$ to $M$}
           \STATE Identify positive prefixes $\mathcal{P}_m$ and negative prefixes $\mathcal{N}_m$
           \STATE Extract latent centroids $\mu_m^+$ and $\mu_m^-$ via Eq. (4)
           \STATE $v_m \leftarrow \mu_m^+ - \mu_m^-$ \COMMENT{Define steering impulse vector}
           \STATE Calculate quality score $q_m$ based on reward gain
           \STATE Store correction entry $\pi = (\mu_m^-, v_m, q_m)$ and anchor entry $(\mu_m^+, 0)$ in $\Omega$
       \ENDFOR
   \ENDFOR
   \STATE // Stage II: Sparse basis selection
   \STATE Select $B$ entries from $\Omega$ by maximizing Eq. (6)
   \STATE // Stage III: Indexing
   \STATE Normalize all keys $k \in \mathcal{M}$ such that $\|k\|_2 = 1$
   \STATE \textbf{return} $\mathcal{M}$
\end{algorithmic}
\end{algorithm}

\begin{algorithm}[htbp]
   \caption{STIR Online: Value-Modulated Trajectory Intervention}
   \label{alg:online}
\begin{algorithmic}[1]
   \REQUIRE Prompt $z$, current prefix $p$, checkpoint index $m$, tool memory $\mathcal{M}$
   \ENSURE Controlled segment or abstention
   \STATE Extract latent descriptor $\phi_{m,l}$ from the residual stream of $p$ at layer $l$
   \STATE $\mathcal{C}_t \leftarrow$ Retrieve top-$k$ entries from $\mathcal{M}$ conditioned on $\phi_{m,l}$
   \STATE $\hat{A}_{anchor} \leftarrow$ Extract maximum similarity score from retrieved anchor entries
   \IF{$\mathcal{C}_t$ is empty \textbf{or} $\hat{A}_{anchor}$ is dominant}
       \STATE \textbf{return} abstain \COMMENT{Consistency-based gating to trust intrinsic generation}
   \ENDIF
   \STATE Filter $\mathcal{C}_t$ to retain top-$L$ candidates based on prior $\hat{A}$
   \FOR{each candidate $\pi_i \in \mathcal{C}_t$}
       \STATE Execute short-horizon preview for $T_{probe}$ tokens using a shared KV-cache
       \STATE Estimate average log-prob gain $\mathcal{G}(\pi_i)$ via Eq. (7)
       \STATE $S(\pi_i) \leftarrow \beta \cdot \hat{A}(\pi_i) + \rho \cdot \mathcal{G}(\pi_i)$ \COMMENT{Compute unified utility score}
   \ENDFOR
   \STATE $\pi^* \leftarrow \arg\max_{\pi_i \in \mathcal{C}_t} S(\pi_i)$
   \IF{$S(\pi^*) < \tau_{null}$}
       \STATE \textbf{return} abstain
   \ELSE
       \STATE $\alpha \leftarrow \min(k_{scale} \cdot S(\pi^*), \alpha_{max})$ \COMMENT{Adaptive strength with clipping}
       \STATE Apply steering impulse $v_{\pi^*}$ to $h_t^{(l)}$ with strength $\alpha$
       \STATE \textbf{return} Generate next segment via $\mathcal{M}_{\theta}(\cdot | h_t^{(l)} + \alpha v_{\pi^*})$
   \ENDIF
\end{algorithmic}
\end{algorithm}

\subsection{Algorithm Logic and Engineering Implementation}
\label{app:pseudocode_and_impl}

The procedural logic of STIR is structured into two distinct phases: the offline construction of the sparse control basis and the online dynamic trajectory modulation.

\paragraph{Offline Memory Construction.}
As outlined in Algorithm~\ref{alg:offline}, the construction phase distills a compact set of steering primitives from noisy model exploration. The process begins by sampling stochastic rollouts for each training example and evaluating them via the length-regularized reward function (Eq.~\ref{eq:reward_function}). At defined structural checkpoints, the algorithm partitions trajectories into positive and negative sets to compute latent centroids (Eq.~\ref{eq:centroids}), deriving raw steering impulses from their vector difference. To manage the redundancy inherent in these raw signals, the final control basis is selected through a diversity-aware optimization process. Specifically, the algorithm employs a greedy determinantal point process~\cite{gillenwater2012near} to iteratively select entries that maximize the marginal gain in the determinantal objective (Eq.~\ref{eq:greedy_selection}). Finally, all state keys undergo L2-normalization to ensure consistency with the angular retrieval metric used during inference.

\paragraph{Online Trajectory Intervention.}
The inference-time execution, detailed in Algorithm~\ref{alg:online}, operates as a retrieve-preview-commit cycle. Upon reaching a decision point, the controller retrieves candidate tools based on angular similarity to the current hidden state and anchor gating is immediately invoked (Lines 4-6). If the retrieved anchor score dominates the candidate set, the system triggers an immediate abstention to preserve the intrinsic optimal trajectory. For the remaining valid candidates, the system executes a short-horizon counterfactual probe to estimate the dynamic reward gain (Eq.~\ref{eq:probe_gain}). This lookahead validation filters out semantically irrelevant retrievals. The system then calculates a unified utility score and enforces a second hard abstention check against the null threshold (Lines 14-15). The steering impulse is injected only if the tool passes this check, with the perturbation magnitude dynamically scaled according to the decision confidence (Eq.~\ref{eq:adaptive_strength}).

\paragraph{Engineering Infrastructure.}
The framework implementation leverages the EasySteer~\cite{xu2025easysteer} library integrated with the vLLM~\cite{kwon2023efficient} engine to support high-throughput generation. The offline discovery phase utilizes data parallelism across GPU nodes to efficiently handle the computational load of large-scale rollout generation. During online inference, the lookahead probing mechanism employs a batched speculative execution strategy. This design allows the system to evaluate multiple candidate tools in parallel within a single forward pass, ensuring that the computational overhead of dynamic validation remains negligible relative to the full reasoning chain.

\begin{figure*}[tb]
    \centering
    \begin{subfigure}{0.24\textwidth}
        \includegraphics[width=\linewidth]{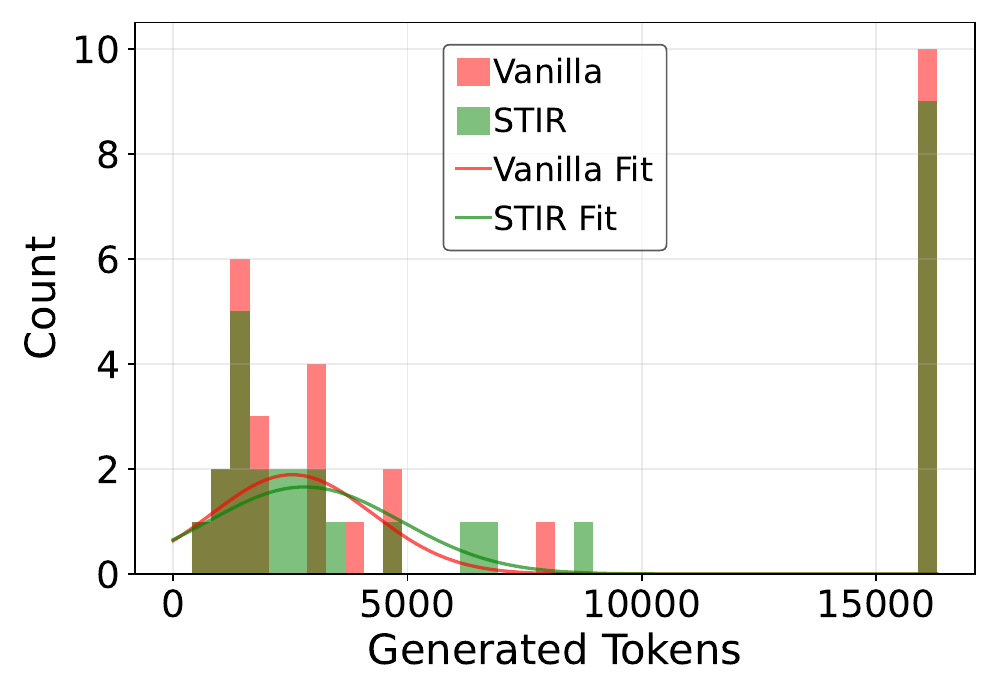}
        \caption{DeepSeek-R1-1.5B ($k{=}0.75$)}
        \label{fig:token_dist_a}
    \end{subfigure}
    \hfill
    \begin{subfigure}{0.24\textwidth}
        \includegraphics[width=\linewidth]{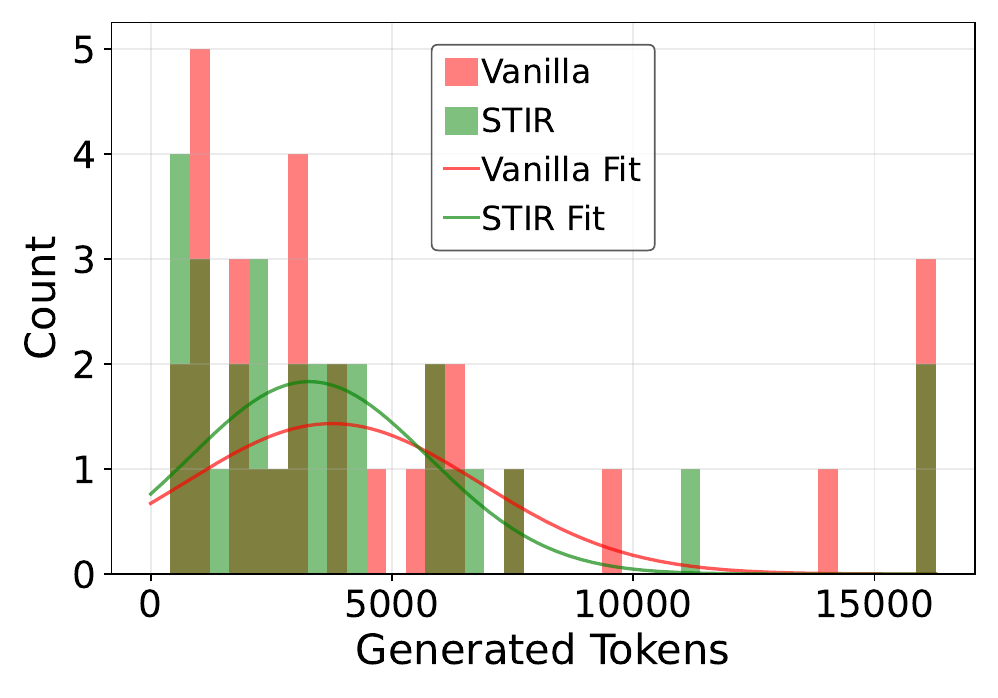}
        \caption{DeepSeek-R1-7B ($k{=}0.75$)}
        \label{fig:token_dist_b}
    \end{subfigure}
    \hfill
    \begin{subfigure}{0.24\textwidth}
        \includegraphics[width=\linewidth]{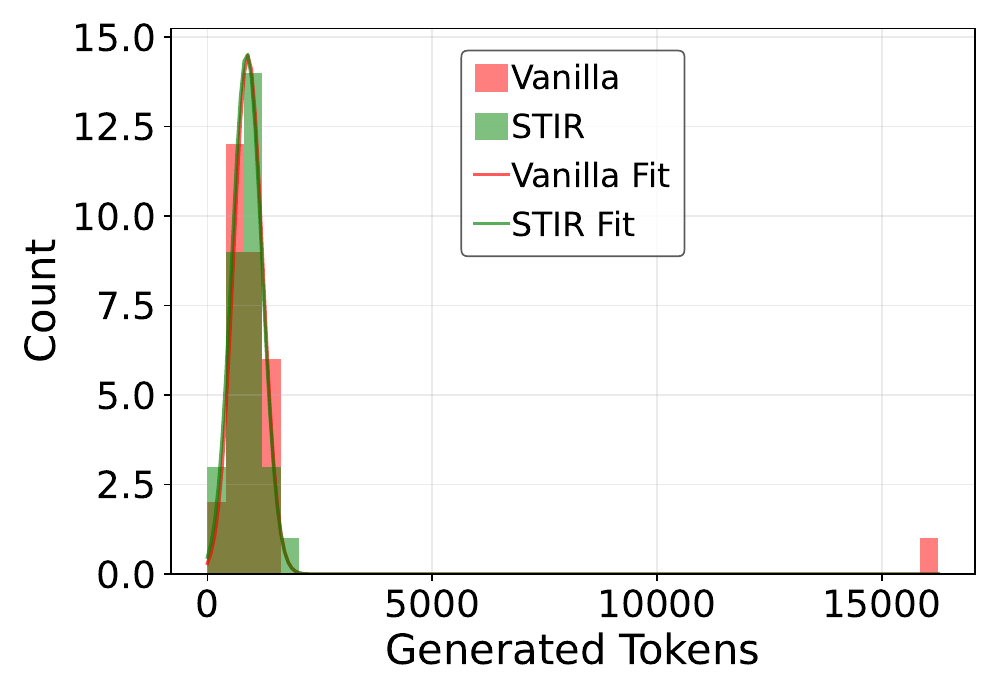}
        \caption{Qwen2.5-3B ($k{=}0.75$)}
        \label{fig:token_dist_c}
    \end{subfigure}
    \hfill
    \begin{subfigure}{0.24\textwidth}
        \includegraphics[width=\linewidth]{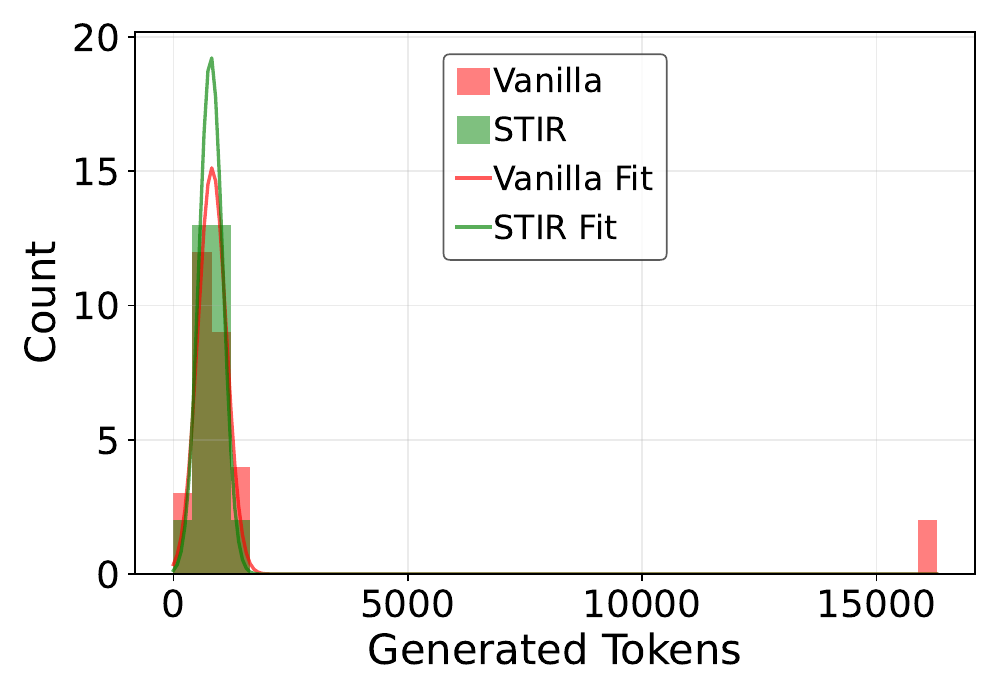}
        \caption{Qwen2.5-7B ($k{=}0.75$)}
        \label{fig:token_dist_d}
    \end{subfigure}

    \par\medskip

    \begin{subfigure}{0.24\textwidth}
        \includegraphics[width=\linewidth]{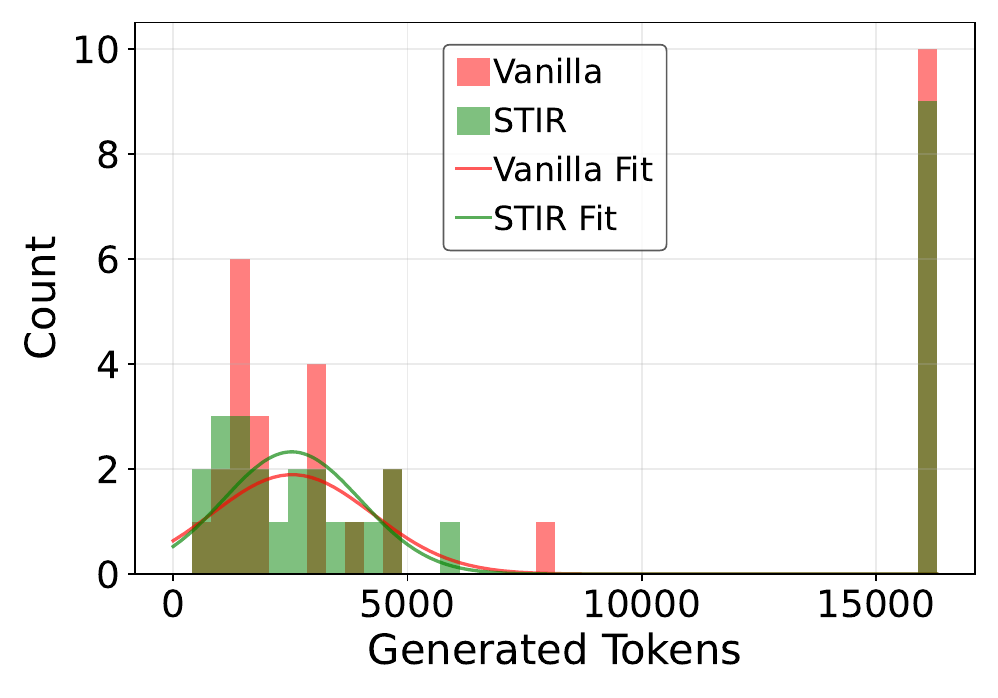}
        \caption{DeepSeek-R1-1.5B ($k{=}1.0$)}
        \label{fig:token_dist_e}
    \end{subfigure}
    \hfill
    \begin{subfigure}{0.24\textwidth}
        \includegraphics[width=\linewidth]{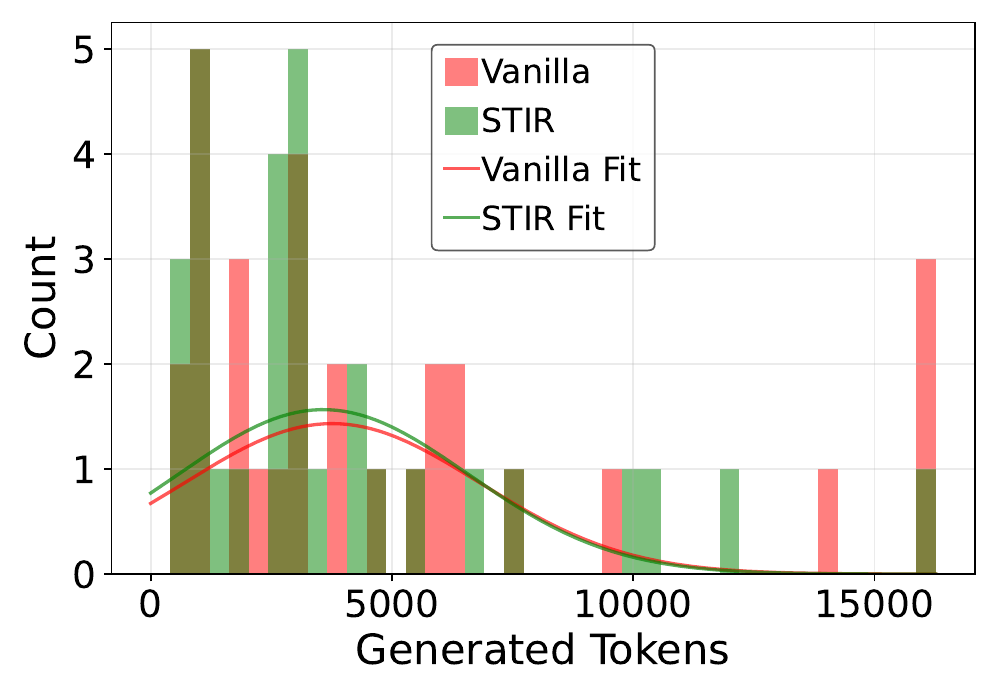}
        \caption{DeepSeek-R1-7B ($k{=}1.0$)}
        \label{fig:token_dist_f}
    \end{subfigure}
    \hfill
    \begin{subfigure}{0.24\textwidth}
        \includegraphics[width=\linewidth]{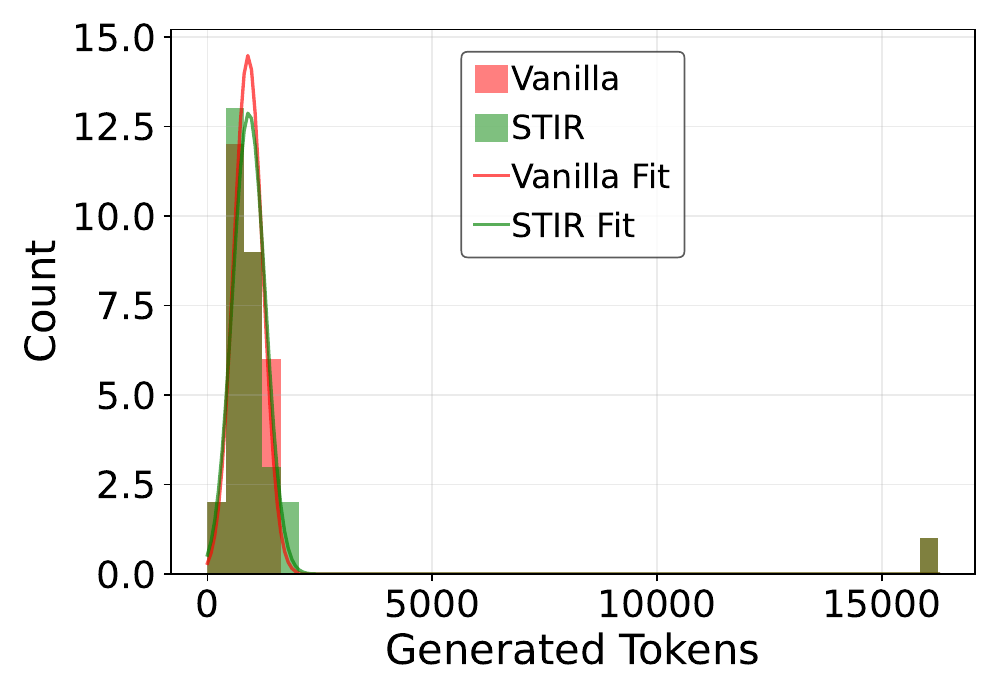}
        \caption{Qwen2.5-3B ($k{=}1.0$)}
        \label{fig:token_dist_g}
    \end{subfigure}
    \hfill
    \begin{subfigure}{0.24\textwidth}
        \includegraphics[width=\linewidth]{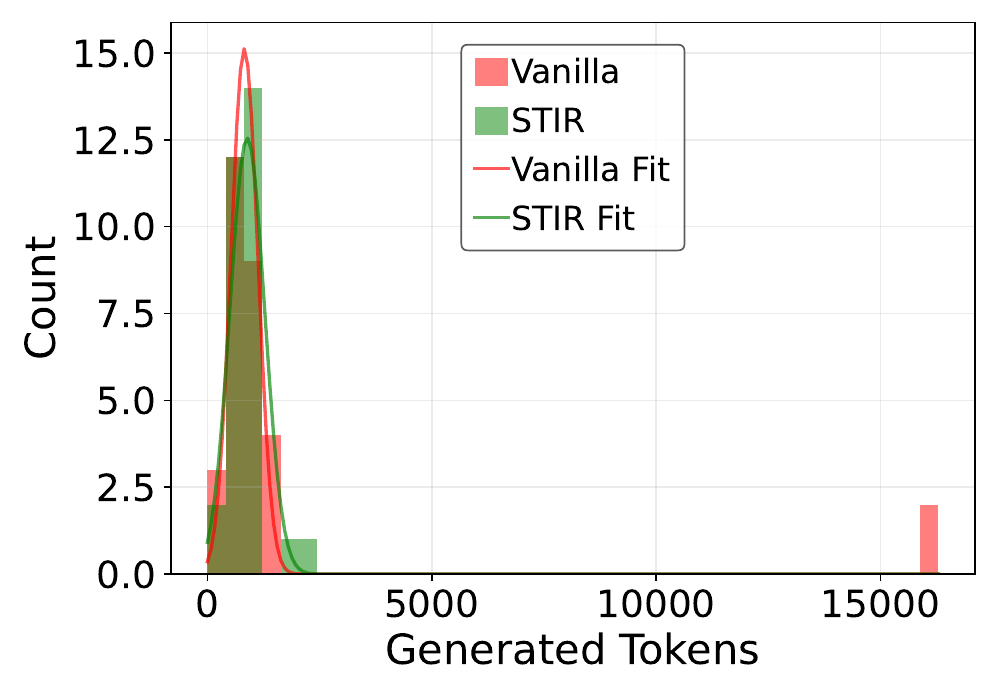}
        \caption{Qwen2.5-7B ($k{=}1.0$)}
        \label{fig:token_dist_h}
    \end{subfigure}

    \caption{Token usage distribution on AMC 23. Comparison between STIR (green) and Vanilla (red) across four models with moderate ($k_{scale}=0.75$, top) and high ($k_{scale}=1.0$, bottom) injection strengths. The uniform leftward shift demonstrates that STIR effectively streamlines reasoning by bypassing redundant steps while simultaneously enhancing solution accuracy.}
    \label{fig:token_distribution}
\end{figure*}

\section{Analysis of Reasoning Length Distribution}
\label{app:length_distribution}

We evaluate the efficiency of STIR by analyzing the distribution of output token lengths on the AMC 23 benchmark. As illustrated in Figure~\ref{fig:token_distribution}, STIR induces a marked leftward shift in token usage compared to the standard chain-of-thought decoding, notably suppressing the long-tail distribution associated with redundant reasoning loops. This compression effect is structurally enforced by the length-regularized reward function employed during the offline discovery phase, which selectively distills steering primitives that favor direct logical progression. Mechanistically, the injected interventions act as computational shortcuts, redirecting the latent state from low-progress manifolds directly toward high-reward conclusion paths. Consequently, STIR reduces average token consumption by 13\%--30\% while simultaneously enhancing solution accuracy, validating that sparse internal control effectively compacts explicit reasoning steps into efficient implicit dynamics.

\begin{MainCaseCard}{Case 1}
    \textbf{Prompt:} Solve the following problem step by step. You should show intermediate steps. Put the final answer in \texttt{\\boxed{<final answer>}} on the last line. Problem: How many zeroes are at the end of $42!$ (42 factorial)?  (Reminder: The number $n!$ is the product of the integers from 1 to $n$.) \hfill \textbf{Gold: 9}
    
    \tcbline
    
    \begin{minipage}[t]{0.48\textwidth}
    \vspace{0pt}
        \begin{tcolorbox}[compare_style, colback=red!2, colframe=red!50!black, title={\ding{55} Vanilla Output (675 Tokens)}]
        \tiny
        
To determine the number of trailing zeros in 42!, I need to count how many times the number 10 is a factor in 42!. Since 10 is the product of 2 and 5, I should find the number of pairs of 2s and 5s in the prime factorization of 42!.

First, I'll count the number of 5s in the factors of 42!. This is done by dividing 42 by 5 and taking the floor of the result. 

42 divided by 5 is 8.4, and taking the floor gives me 8. So, there are 8 factors of 5 in 42!.

Next, I'll count the number of 2s. I'll divide 42 by 2, then by 4, 8, 16, and 32, and take the floor of each result and sum them up.

42 divided by 2 is 21, by 4 is 10, by 8 is 5, by 16 is 2, and by 32 is 1. Adding these together: 21 + 10 + 5 + 2 + 1 equals 40. So, there are 40 factors of 2 in 42!.

Since there are more 2s than 5s, the number of trailing zeros is determined by the number of 5s, which is 8.
\texttt{</think>}

To determine how many trailing zeros are in \(42!\), we need to count the number of times 10 is a factor in the product. Since \(10 = 2 \times 5\), we count the number of pairs of 2s and 5s in the prime factorization of \(42!\). Typically, there are more 2s than 5s, so the number of 5s will determine the number of trailing zeros.

**Step 1: Count the number of 5s in the factors of \(42!\).**

\[
\left\lfloor \frac{42}{5} \right\rfloor = 8
\]

So, there are 8 factors of 5.

**Step 2: Count the number of 2s in the factors of \(42!\).**

\[
\left\lfloor \frac{42}{2} \right\rfloor + \left\lfloor \frac{42}{4} \right\rfloor + \left\lfloor \frac{42}{8} \right\rfloor + \left\lfloor \frac{42}{16} \right\rfloor + \left\lfloor \frac{42}{32} \right\rfloor = 21 + 10 + 5 + 2 + 1 = 40
\]

There are 40 factors of 2.

**Step 3: Determine the number of trailing zeros.**

Since there are more 2s than 5s, the number of trailing zeros is equal to the number of 5s.

\[
\boxed{8}
\]
        \end{tcolorbox}
    \end{minipage}
    \hfill
    \begin{minipage}[t]{0.48\textwidth}
    \vspace{0pt}
        \begin{tcolorbox}[compare_style, colback=green!2, colframe=green!50!black, title={\ding{51} STIR Output (496 Tokens)}]
        \tiny
1. **Understand the Problem:**  
   I need to find the number of trailing zeros in 42!.

2. **Recall the Method:**  
   Trailing zeros in a factorial are created by multiplying 10s, which come from pairs of 2 and 5. Since there are usually more 2s than 5s, the number of 5s determines the number of trailing zeros.

3. **Calculate the Number of 5s:**
   - Divide 42 by 5: 42 ÷ 5 = 8.4, so there are 8 multiples of 5.
   - Divide 42 by 25: 42 ÷ 25 = 1.68, so there is 1 multiple of 25.

4. **Sum the Results:**  
   8 (from multiples of 5) + 1 (from multiples of 25) = 9.

5. **Conclusion:**  
   There are 9 trailing zeros in 42!.
\texttt{</think>}

 **Solution:**

 To determine the number of trailing zeros in \(42!\), we need to count the number of times 10 is a factor in the product. Since \(10 = 2 \times 5\), and there are usually more factors of 2 than 5 in a factorial, the number of trailing zeros is determined by the number of times 5 is a factor in the numbers from 1 to 42.

**Step 1: Count the multiples of 5**

\[
\left\lfloor \frac{42}{5} \right\rfloor = 8
\]

This accounts for the numbers 5, 10, 15, 20, 25, 30, 35, and 40.

**Step 2: Count the multiples of 25 (since 25 contributes an extra factor of 5)**

\[
\left\lfloor \frac{42}{25} \right\rfloor = 1
\]

This accounts for the number 25, which contributes an additional factor of 5.

**Step 3: Sum the counts**

\[
8 + 1 = 9
\]

**Final Answer:**

\[
\boxed{9}
\]
        \end{tcolorbox}
    \end{minipage}
\end{MainCaseCard}

\section{Case Studies}
\label{app:case_studies}

We analyze the reasoning trajectories of DeepSeek-R1-Distill-Qwen-7B on the MATH500 benchmark to show how dynamic steering corrects the failures of chain-of-thought decoding. In number theory problems requiring factorial decomposition, STIR retrieves steering signals that explicitly activate the concept of higher-order prime powers, effectively filling in calculation steps that the vanilla model missed. Similarly, in algebraic tasks involving non-linear equations, the anchor-based gating mechanism identifies high-uncertainty transitions and prevents the model from incorrectly discarding valid solutions during verification. These examples demonstrate that the framework can both supply missing algorithmic steps and improve logical consistency, validating the effectiveness of sparse internal control in restoring correct reasoning paths.

\section{Limitations}
\label{app:limitations}

Despite the empirical gains demonstrated by STIR, several limitations inherent to the current methodology require further investigation. The framework relies on extracting steering impulses at predefined structural delimiters, which may restrict the granularity of trajectory control and potentially overlook reasoning slips that occur within dense logical segments. Additionally, the corrective efficacy of the tool library is bounded by the geometric diversity of failure modes captured during the offline discovery phase, meaning that rare logical fallacies that do not manifest during the intrinsic contrastive sampling may remain unaddressable by the sparse control basis. Furthermore, the effectiveness of the anchor based gating mechanism depends heavily on the precision of latent similarity metrics in high-dimensional activation spaces, where distinguishing deviations from valid stylistic variations remains challenging in regimes with narrow margins between optimal and erroneous manifolds. Finally, our current evaluation primarily targets verification-friendly reasoning tasks, and extending this internal control paradigm to open-ended domains with subjective correctness constitutes an open challenge for future research.

\section{Broader Impact}
\label{app:broader_impact}

This work introduces STIR to enhance the reliability and efficiency of Large Language Models by internalizing explicit reasoning into dynamic latent trajectory control. By decoupling reasoning depth from generation length, our framework significantly advances computational sustainability and facilitates the deployment of trustworthy AI in resource-constrained critical domains. While latent activation steering inherently entails dual use risks regarding the potential induction of biased or harmful behaviours, STIR mitigates these ethical concerns through its self-distilled and evidence-driven design by strictly limiting interventions to intrinsically verified high reward trajectories. Consequently, STIR establishes a robust paradigm for safer, more interpretable, and resource-efficient autonomous reasoning systems, thereby reducing the attack surface for adversarial manipulation.

\newpage
\begin{MainCaseCard}{Case 2}
    \textbf{Prompt:} Solve the following problem step by step. You should show intermediate steps. Put the final answer in \texttt{\\boxed{<final answer>}} on the last line. Problem: If $|x+5|-|3x-6|=0$, find the largest possible value of $x$. Express your answer as an improper fraction. \hfill \textbf{Gold: 11/2}
    
    \tcbline
    
    \begin{minipage}[t]{0.48\textwidth}
    \vspace{0pt}
        \begin{tcolorbox}[compare_style, colback=red!2, colframe=red!50!black, title={\ding{55} Vanilla Output (1325 Tokens)}]
        \tiny
To solve the equation $|x + 5| - |3x - 6| = 0$, I need to find the values of x that satisfy this condition.

First, I'll set up the equation by moving one absolute value term to the other side:
$|x + 5| = |3x - 6|$

Next, I'll consider the definition of absolute value, which leads to two possible cases for each side. This results in four different equations to solve.

For each case, I'll solve for x and then check if the solution is valid by substituting it back into the original equation.

After finding all possible solutions, I'll identify the largest value of x among them.
\texttt{</think>}

To solve the equation \( |x + 5| - |3x - 6| = 0 \) and find the largest possible value of \( x \), follow these steps:

\#\#\# Step 1: Set Up the Equation
Start by isolating one of the absolute value expressions:
\[
|x + 5| = |3x - 6|
\]

\#\#\# Step 2: Consider the Cases for Absolute Values
The equation \( |A| = |B| \) implies two possible cases:
1. \( A = B \)
2. \( A = -B \)

Apply these cases to the equation:

**Case 1:** \( x + 5 = 3x - 6 \)
\begin{align*}
x + 5 &= 3x - 6 \\
5 + 6 &= 3x - x \\
11 &= 2x \\
x &= \frac{11}{2}
\end{align*}

**Case 2:** \( x + 5 = -(3x - 6) \)
\begin{align*}
x + 5 &= -3x + 6 \\
x + 3x &= 6 - 5 \\
4x &= 1 \\
x &= \frac{1}{4}
\end{align*}

\#\#\# Step 3: Verify the Solutions
It's essential to check each solution in the original equation to ensure it's valid.

**For \( x = \frac{11}{2} \):**
\begin{align*}
\left| \frac{11}{2} + 5 \right| - \left| 3 \cdot \frac{11}{2} - 6 \right| &= \left| \frac{21}{2} \right| - \left| \frac{21}{2} - 6 \right| \\
&= \frac{21}{2} - \frac{9}{2} = \frac{12}{2} = 6 \neq 0
\end{align*}
This solution does not satisfy the original equation.

**For \( x = \frac{1}{4} \):**
\begin{align*}
\left| \frac{1}{4} + 5 \right| - \left| 3 \cdot \frac{1}{4} - 6 \right| &= \left| \frac{21}{4} \right| - \left| \frac{3}{4} - 6 \right| \\
&= \frac{21}{4} - \left| -\frac{21}{4} \right| = \frac{21}{4} - \frac{21}{4} = 0
\end{align*}
This solution satisfies the original equation.

\#\#\# Step 4: Identify the Largest Valid Solution
From the verified solutions, only \( x = \frac{1}{4} \) is valid. However, upon re-examining the problem, it's clear that there might be another valid solution when considering the absolute values more carefully.

**Re-examining Case 1:**
\[
x + 5 = 3x - 6 \implies x = \frac{11}{2}
\]
But substituting \( x = \frac{11}{2} \) back into the original equation:
\[
\left| \frac{11}{2} + 5 \right| - \left| 3 \cdot \frac{11}{2} - 6 \right| = \frac{21}{2} - \frac{9}{2} = 6 \neq 0
\]
This indicates that \( x = \frac{11}{2} \) is not a valid solution.

**Considering Another Approach:**
Let's analyze the equation by squaring both sides to eliminate the absolute values:
\[
|x + 5| = |3x - 6| \implies (x + 5)^2 = (3x - 6)^2
\]
Expanding both sides:
\[
x^2 + 10x + 25 = 9x^2 - 36x + 36
\]
Bringing all terms to one side:
\[
0 = 8x^2 - 46x + 11
\]
Solving the quadratic equation:
\begin{align*}
x = \frac{46 \pm \sqrt{(-46)^2 - 4 \cdot 8 \cdot 11}}{2 \cdot 8} = \frac{46 \pm \sqrt{2116 - 352}}{16} \\
= \frac{46 \pm \sqrt{1764}}{16} = \frac{46 \pm 42}{16}
\end{align*}
Thus, the solutions are:
\[
x = \frac{88}{16} = \frac{11}{2} \quad \text{and} \quad x = \frac{4}{16} = \frac{1}{4}
\]
As previously determined, \( x = \frac{11}{2} \) does not satisfy the original equation, leaving \( x = \frac{1}{4} \) as the only valid solution.

\#\#\# Final Answer
\[
\boxed{\dfrac{1}{4}}
\]       
        \end{tcolorbox}
    \end{minipage}
    \hfill
    \begin{minipage}[t]{0.48\textwidth}
    \vspace{0pt}
        \begin{tcolorbox}[compare_style, colback=green!2, colframe=green!50!black, title={\ding{51} STIR Output (1122 Tokens)}]
        \tiny
To solve the equation $|x + 5| - |3x - 6| = 0$, I need to find the values of x that satisfy this condition.

First, I'll set up the equation by isolating one of the absolute value expressions:
$|x + 5| = |3x - 6|$

Next, I'll consider the different cases based on the critical points where the expressions inside the absolute values change their signs. The critical points are x = -5 and x = 2.

I'll analyze each interval determined by these critical points:

1. **Case 1: x < -5**
   - Both expressions inside the absolute values are negative.
   - The equation becomes -(x + 5) = -(3x - 6).
   - Solving this gives x = 1/2, which is not in the interval x < -5.

2. **Case 2: -5 $\le$ x < 2**
   - x + 5 is non-negative, and 3x - 6 is negative.
   - The equation becomes (x + 5) = -(3x - 6).
   - Solving this gives x = 1/4, which is within the interval -5 $\le$ x < 2.

3. **Case 3: x $\ge$ 2**
   - Both expressions inside the absolute values are non-negative.
   - The equation becomes (x + 5) = (3x - 6).
   - Solving this gives x = 11/2, which is within the interval x $\ge$ 2.

4. **Case 4: x < -5**
   - This case was already covered in Case 1 and did not yield a valid solution.

5. **Case 5: -5 $\le$ x < 2**
   - This case was covered in Case 2 and yielded x = 1/4.

6. **Case 6: x $\ge$ 2**
   - This case was covered in Case 3 and yielded x = 11/2.

Finally, I'll compare the valid solutions x = 1/4 and x = 11/2 to determine the largest possible value of x, which is 11/2.
\texttt{</think>}

To solve the equation \( |x + 5| - |3x - 6| = 0 \) and find the largest possible value of \( x \), follow these steps:

\#\#\# Step 1: Set Up the Equation
Start by isolating one of the absolute value expressions:
\[
|x + 5| = |3x - 6|
\]

\#\#\# Step 2: Identify Critical Points
The expressions inside the absolute values change their signs at:
\[
x + 5 = 0 \quad \Rightarrow \quad x = -5
\]
\[
3x - 6 = 0 \quad \Rightarrow \quad x = 2
\]
These points divide the real number line into three intervals:
1. \( x < -5 \)  
2. \( -5 \leq x < 2 \)  
3. \( x \geq 2 \)

\#\#\# Step 3: Solve in Each Interval

\#\#\#\# **Case 1: \( x < -5 \)**
Both \( x + 5 \) and \( 3x - 6 \) are negative.
\[
-(x + 5) = -(3x - 6)
\]
\[
-x - 5 = -3x + 6
\]
\[
2x = 11 \quad \Rightarrow \quad x = \frac{11}{2}
\]
Since \( \frac{11}{2} = 5.5 \) is not less than -5, there's no solution in this interval.

\#\#\#\# **Case 2: \( -5 \leq x < 2 \)**
\( x + 5 \) is non-negative, and \( 3x - 6 \) is negative.
\[
(x + 5) = -(3x - 6)
\]
\[
x + 5 = -3x + 6
\]
\[
4x = 1 \quad \Rightarrow \quad x = \frac{1}{4}
\]
\( \frac{1}{4} \) is within the interval \( -5 \leq x < 2 \), so this is a valid solution.

\#\#\#\# **Case 3: \( x \geq 2 \)**
Both \( x + 5 \) and \( 3x - 6 \) are non-negative.
\[
(x + 5) = (3x - 6)
\]
\[
2x = 11 \quad \Rightarrow \quad x = \frac{11}{2}
\]
\( \frac{11}{2} = 5.5 \) is greater than or equal to 2, so this is a valid solution.

\#\#\# Step 4: Compare Solutions
The valid solutions are:
\[
x = \frac{1}{4} \quad \text{and} \quad x = \frac{11}{2}
\]
The largest possible value of \( x \) is \( \frac{11}{2} \).

\#\#\# Final Answer
\[
\boxed{\dfrac{11}{2}}
\]
        \end{tcolorbox}
    \end{minipage}
\end{MainCaseCard}

\end{document}